\pdfoutput=1
\documentclass[11pt]{article}
\usepackage[margin=1in]{geometry}
\usepackage{amsmath,amssymb,amsthm,mathtools}
\usepackage{bm}
\usepackage[numbers,sort&compress]{natbib}
\usepackage[colorlinks=true,linkcolor=blue,citecolor=blue,urlcolor=blue]{hyperref}
\usepackage{microtype}
\usepackage{graphicx}
\usepackage{booktabs}
\usepackage{xcolor}
\usepackage{tikz}
\usetikzlibrary{arrows.meta,calc,fit,backgrounds,positioning}
\usepackage{pgfplots}
\pgfplotsset{compat=1.18}
\usepackage{algorithm}
\usepackage[section]{placeins}
\usepackage{algpseudocode}

\definecolor{figblue}{RGB}{40,55,190}
\definecolor{figgreen}{RGB}{0,120,60}
\definecolor{figred}{RGB}{190,40,40}
\definecolor{figgray}{RGB}{130,130,130}

\tikzset{
  taskvec/.style  = {figblue, -{Stealth[length=2.4mm]}, thick},
  dispvec/.style  = {figgreen, -{Stealth[length=2.4mm]}, very thick},
  aux/.style      = {figgray, dashed, thin},
  budget/.style   = {figblue, thick, dashed},
  traj/.style     = {figgreen, very thick},
  levelset/.style = {black, thin},
  pt/.style       = {circle, fill=black, inner sep=1.35pt},
  ptopen/.style   = {circle, draw=black, fill=white, inner sep=1.2pt},
  lbl/.style      = {font=\footnotesize},
  lblb/.style     = {font=\footnotesize, figblue},
  lblg/.style     = {font=\footnotesize, figgreen},
}

\algrenewcommand\algorithmicrequire{\textbf{Input:}}
\algrenewcommand\algorithmicensure{\textbf{Output:}}
\newcommand{\algcomment}[1]{\hfill\textcolor{figgray}{\footnotesize$\triangleright$ #1}}

\newtheorem{theorem}{Theorem}[section]
\newtheorem{lemma}[theorem]{Lemma}
\newtheorem{corollary}[theorem]{Corollary}
\newtheorem{proposition}[theorem]{Proposition}
\theoremstyle{definition}
\newtheorem{definition}[theorem]{Definition}
\newtheorem{assumption}{Assumption}
\newtheorem{example}[theorem]{Example}
\theoremstyle{remark}
\newtheorem{remark}[theorem]{Remark}

\newcommand{\R}{\mathbb{R}}
\newcommand{\E}{\mathbb{E}}
\newcommand{\loss}{\mathcal{L}}
\newcommand{\Hess}{H}
\newcommand{\sal}{\bar{S}}
\newcommand{\merged}{\theta(\bm{\lambda})}
\newcommand{\Ov}{\Omega}
\newcommand{\norm}[1]{\left\lVert #1 \right\rVert}

\title{Interference Bounds and Merge-Aware Training Dynamics\\
for Composable Fine-Tuned Models}
\author{Alireza Moayedikia\thanks{Alireza Moayedikia is with the
Department of Business Technology and Entrepreneurship, Swinburne
University of Technology, John Street, Hawthorn, VIC 3122, Australia
(e-mail: amoayedikia@swin.edu.au). (Corresponding author:
Alireza Moayedikia.)}\\[4pt]
\normalsize Department of Business Technology and Entrepreneurship\\
\normalsize Swinburne University of Technology}
\date{}

\makeatletter
\renewcommand{\@maketitle}{%
  \newpage\null\vskip 2em%
  \begin{center}{\LARGE \@title \par}\end{center}%
  \vskip 1.5em%
  {\large \raggedright \@author \par}%
  \vskip 1em%
  \begin{center}\@date\end{center}%
  \par\vskip 1.5em}
\makeatother

\begin{document}
\maketitle

\begin{abstract}
Combining fine-tuned models by composing their parameter updates has become a standard alternative to multi-task training. We formalize composability as a property of a collection of task vectors and prove that the excess loss of the merged model is controlled by an interference functional of the cross-task curvature overlap and the signed alignment of the updates. In the quadratic regime this functional equals the excess loss, giving a criterion that is necessary as well as sufficient and that measures the angle between task vectors in the curvature metric rather than the Euclidean one. Fine-tuning under a curvature-guided regularizer---constructed from curvature statistics of the kind memory-efficient optimizers maintain---keeps displacement in the base curvature metric below a ceiling set by the penalty coefficient at every iterate, at the standard $O(1/\sqrt{T})$ non-convex rate. Under a single Loewner comparison between the base statistic and the task curvatures, this control converts into a ceiling on the interference functional, inversely proportional to the penalty coefficient, so that the curvature term of the bound is controlled during training. A first-order term, present because regularized solutions are not stationary for the task loss, is not reduced in the same way. Experiments on an exact non-quadratic model confirm the budget and the ceiling, show the first-order term dominating for strong penalties, and find the curvature-guided penalty slightly better than an isotropic one. In a three-seed transformer study the functional is rank-correlated with merging degradation, but no more strongly than the Euclidean displacement norm.
\end{abstract}

\vspace{0.6em}

\section{Introduction}\label{sec:intro}

Fine-tuning a shared pre-trained model $\theta_0 \in \R^d$ on $K$ separate
tasks produces parameters $\theta_k = \theta_0 + \tau_k$, where
$\tau_k \in \R^d$ is the \emph{task vector} of task $k$
\citep{ilharco2023editing}. The practice of \emph{merging}---forming
$\merged = \theta_0 + \sum_{k=1}^{K} \lambda_k \tau_k$ for weights
$\lambda_k \ge 0$---promises multi-task capability without joint training,
and a rapidly growing family of operators---from plain weight
averaging \citep{izmailov2018averaging, wortsman2022soups} to
Fisher-weighted averaging \citep{matena2022fisher}, sign-consistent
sparsification \citep{yadav2023ties}, random dropping and rescaling
\citep{yu2024dare}, and alignment-based re-basin
\citep{ainsworth2023gitrebasin}---attempts to suppress the
\emph{interference} between task vectors that degrades merged
performance, echoing the gradient-interference problem long studied in
multi-task learning \citep{yu2020gradient}. These operators are heuristics: each
implicitly targets a different source of interference, each fails on some
task collections, and the field currently lacks a criterion that says,
before merging, whether a given collection of fine-tuned models can be
composed at all.

This paper develops such a criterion and shows it can be enforced during
training. Our starting point is the observation, developed empirically in
our earlier work \citep{moayedikia2026bridging}, that the curvature
information adaptive optimizers accumulate during fine-tuning is precisely
the information that determines merging quality. Here we give this
observation its mathematical form. We ask two questions. First: for which
collections $\{\tau_k\}$ and weights $\bm{\lambda}$ does the merged model
provably retain each task's performance? Second: can the fine-tuning
dynamics be modified---at negligible cost---so that the resulting task
vectors satisfy the sufficient conditions by construction?

Our contributions are as follows. We introduce
$\varepsilon$-composability, a formal property of a collection of task
vectors relative to their task losses
(Definition~\ref{def:composability}). We prove an interference bound
(Theorem~\ref{thm:bound}): the merged model's excess loss on task $k$ is
controlled by a functional $\Ov_k$ of the cross-task curvature overlap and
the alignment structure of the task vectors, plus a cubic remainder; the
decomposition suggests a heuristic reading of Fisher-weighted averaging and sign election as acting on distinct terms. In the quadratic (linearized) regime we
prove the criterion is exact (Theorem~\ref{thm:characterization}):
$\varepsilon$-composability holds if and only if the interference
functional is at most $2\varepsilon$, and the bound is attained by an
explicit two-task construction. We then analyze fine-tuning under a
curvature-guided proximity regularizer and prove
(Theorem~\ref{thm:dynamics}) that the regularized dynamics converge at the
standard non-convex rate while keeping displacement in the base curvature
metric below a ceiling set by the penalty coefficient at every iterate.
Because that ceiling is stated in the base metric while the interference
functional is stated in the task curvatures, we then isolate the single
comparison between the two that closes the gap
(Assumption~\ref{ass:domination}) and prove
(Theorem~\ref{thm:ceiling}) that training each task separately in this
way bounds the interference functional of the merged collection by a
quantity proportional to $1/\mu$ and to the task-loss descent each run
achieves---a merge-time certificate assembled entirely from single-task
training outputs and the comparison constant $c$, with no access to the siblings. We also identify (Proposition~\ref{prop:firstorder}) the term this ceiling does not control: a regularized solution is stationary for the penalized objective rather than for the task loss, so the excess loss of the merged model acquires a first-order term governed by the alignment of the task vectors in the base metric. Finally we show (Corollary~\ref{cor:factorized}) that
all guarantees survive, with graceful degradation, when the curvature
statistic is maintained in factorized low-rank form as done by
memory-efficient optimizers, with error controlled by the spectral
tail of the curvature statistic.

We state plainly what kind of contribution this is. The individual arguments are elementary---a second-order expansion, a sublevel-set rearrangement, a Loewner comparison and a triangle inequality---and we claim no technical novelty for any of them. The contribution is the formulation: a single functional that organizes the merging problem, an explicit statement of which quantity each training-time guarantee controls and under which assumption, and an equally explicit account (Proposition~\ref{prop:firstorder}, Remark~\ref{rem:isotropic}) of what the guarantees do not deliver.

The paper is organized as follows. Section~\ref{sec:setting} fixes the
setting and assumptions. Section~\ref{sec:bound} states the interference
bound, Section~\ref{sec:characterization} the exact characterization, and
Section~\ref{sec:dynamics} the training-dynamics results.
Section~\ref{sec:numerics} reports numerical experiments.
Section~\ref{sec:related} positions the results against existing theory.
Proofs are deferred to the appendix.

\paragraph{Notation.} $\norm{\cdot}$ denotes the Euclidean norm and, for a
symmetric positive semi-definite matrix $A$, $\norm{x}_A^2 = x^\top A x$.
For symmetric matrices, $\preceq$ denotes the Loewner order.
$\sigma_i(M)$ denotes the $i$-th singular value of $M$ in decreasing
order. $\odot$ denotes the Hadamard product. $[K] = \{1,\dots,K\}$.

\section{Setting and Assumptions}\label{sec:setting}

Each task $k \in [K]$ has a data distribution $\mathcal{D}_k$ and
population loss $\loss_k(\theta) = \E_{\xi \sim \mathcal{D}_k}
[\ell(\theta,\xi)]$. Fine-tuning task $k$ from the shared initialization
$\theta_0$ returns $\theta_k$, assumed to be an approximate local minimizer
of $\loss_k$, and we write $\tau_k = \theta_k - \theta_0$. Given merging
weights $\bm{\lambda} = (\lambda_1,\dots,\lambda_K)$, the merged model is
\begin{equation}\label{eq:merged}
\merged \;=\; \theta_0 + \sum_{k=1}^{K} \lambda_k \tau_k .
\end{equation}
The displacement of the merged model from the task-$k$ solution is
\begin{equation}\label{eq:displacement}
d_k(\bm{\lambda}) \;=\; \merged - \theta_k
\;=\; (\lambda_k - 1)\,\tau_k + \sum_{j \neq k} \lambda_j \tau_j .
\end{equation}
Throughout, $\Hess_k = \nabla^2 \loss_k(\theta_k)$ denotes the task-$k$
Hessian at the fine-tuned solution. Table~\ref{tab:notation} collects
the notation used throughout the paper for reference.

\begin{table}[t]
\centering
\small
\setlength{\tabcolsep}{4pt}
\renewcommand{\arraystretch}{1.1}
\caption{Summary of notation.}
\label{tab:notation}
\begin{tabular}{@{}l@{\ }p{0.305\linewidth}@{\hspace{1.2em}}l@{\ }p{0.305\linewidth}@{}}
\toprule
\multicolumn{4}{@{}l}{\textit{Models and merging}}\\
\midrule
$d$ & parameter dimension & $\bm{\lambda}$ & merging weights $(\lambda_1,\dots,\lambda_K)$\\
$K,[K]$ & \# tasks; $\{1,\dots,K\}$ & $\Lambda$ & admissible weight set\\
$\theta_0$ & base (pre-fine-tuning) model & $\merged$ & merged model, Eq.~\eqref{eq:merged}\\
$\theta_k$ & task-$k$ fine-tuned model & $d_k(\bm{\lambda})$ & displacement $\merged-\theta_k$, Eq.~\eqref{eq:displacement}\\
$\tau_k$ & task vector $\theta_k-\theta_0$ & &\\
\addlinespace[2pt]
\multicolumn{4}{@{}l}{\textit{Losses and curvature}}\\
\midrule
$\mathcal{D}_k$ & task-$k$ data distribution & $\Hess_k$ & task-$k$ Hessian at $\theta_k$\\
$\ell(\theta,\xi)$ & per-example loss & $\Ov_k(\bm{\lambda})$ & interference functional, Eq.~\eqref{eq:omega}\\
$\loss_k$ & task-$k$ population loss & $\varepsilon$ & composability tolerance (Def.~\ref{def:composability})\\
$\loss_k^\star$ & minimum of $\loss_k$ & $c$ & curvature domination constant (Assum.~\ref{ass:domination})\\
\addlinespace[2pt]
\multicolumn{4}{@{}l}{\textit{Saliency statistics}}\\
\midrule
$\sal_k$ & task-$k$ saliency (byproduct) & $\widehat{\Ov}_k$ & interference computed from $\sal_k$\\
$\sal_0$ & base-model saliency & $\epsilon_r(\sal)$ & truncation error $\sigma_{r+1}(\sal)$\\
$\sal^{(r)}$ & rank-$r$ factorized approx.\ of $\sal$ & $\delta$ & saliency approx.\ gap (Assum.~\ref{ass:saliency})\\
$r,\eta$ & effective-rank params (Assum.~\ref{ass:rank}) & &\\
\addlinespace[2pt]
\multicolumn{4}{@{}l}{\textit{Optimization}}\\
\midrule
$\mu$ & penalty coefficient, Eq.~\eqref{eq:regularized} & $\rho$ & Hessian-Lipschitz constant\\
$J_k$ & regularized objective, Eq.~\eqref{eq:regularized} & $L_\mu$ & smoothness constant of $J_k$\\
$\Phi(\theta)$ & surrogate interference & $\epsilon_g$ & stationarity slack (Assum.~\ref{ass:stationary})\\
$L$ & smoothness constant of $\loss_k$ & $\theta^{(t)}$ & iterate at step $t$\\
$\gamma,\gamma_t$ & step size & $T$ & number of iterations\\
$\sigma^2$ & gradient-noise variance & $\Delta_k$ & task-loss descent $\loss_k(\theta_0)-\loss_k(\hat\theta_k)$ (Thm.~\ref{thm:ceiling})\\
\addlinespace[2pt]
\multicolumn{4}{@{}l}{\textit{General}}\\
\midrule
$\norm{\cdot}$ & Euclidean norm & $\sigma_i(M)$ & $i$-th singular value (decr.)\\
$\norm{x}_A^2$ & $x^\top A x$, for $A\succeq0$ & $\preceq$ & Loewner order\\
$\norm{\cdot}_{\mathrm{op}},\norm{\cdot}_*$ & operator, trace norm & $\odot$ & Hadamard product\\
$\mathcal{N}$ & neighborhood (Assum.~\ref{ass:smooth}) & &\\
\bottomrule
\end{tabular}
\end{table}

\begin{assumption}[Smoothness]\label{ass:smooth}
Let $\Lambda \subset \R_{\ge 0}^K$ denote the set of merging weights
under consideration. Each $\loss_k$ is twice continuously
differentiable, $L$-smooth, and has $\rho$-Lipschitz Hessian on a
convex neighborhood $\mathcal{N}$ containing
$\{\theta_0\} \cup \{\theta_k\}_{k \in [K]} \cup
\{\merged : \bm{\lambda} \in \Lambda\}$.
\end{assumption}

Assumption~\ref{ass:smooth} is standard in non-convex optimization and is
what licenses the second-order expansion with controlled cubic remainder;
the Hessian-Lipschitz constant $\rho$ enters only the remainder term.

\begin{assumption}[Approximate stationarity]\label{ass:stationary}
There exists $\epsilon_g \ge 0$ such that
$\norm{\nabla \loss_k(\theta_k)} \le \epsilon_g$ for all $k \in [K]$.
\end{assumption}

Fine-tuned models are not exact minimizers; $\epsilon_g$ carries this
slack through the bounds. Setting $\epsilon_g = 0$ recovers the idealized statements. The slack is not always small: when $\theta_k$ is produced by the regularized training of Section~\ref{sec:dynamics} it is stationary for the penalized objective $J_k$ rather than for $\loss_k$, and $\nabla\loss_k(\theta_k) = -\mu\,\sal_0\tau_k$ there. Proposition~\ref{prop:firstorder} treats this case explicitly instead of absorbing it into $\epsilon_g$.

\begin{assumption}[Low effective curvature rank]\label{ass:rank}
There exist $r \ll d$ and $\eta \in [0,1)$ such that for each $k$,
$\sum_{i > r} \sigma_i(\Hess_k) \le \eta \sum_{i \ge 1} \sigma_i(\Hess_k)$.
\end{assumption}

Assumption~\ref{ass:rank} is motivated by empirical observations on trained networks. A substantial line of work
documents that the Hessian spectrum of trained networks splits into a
small number of outliers and a bulk concentrated near zero
\citep{sagun2018empirical, ghorbani2019investigation, papyan2020traces},
and that the gradient concentrates in the corresponding low-dimensional
top-eigenvector subspace \citep{gurari2018gradient}; spectral
measurements in \citep{moayedikia2026bridging} confirm the same for the
trajectory statistics used here, with stable rank near $3$ and over
$82\%$ of the energy captured at rank $32$. The assumption is used only
in Section~\ref{sec:dynamics}, where it controls the factorization error.

\begin{assumption}[Saliency approximation quality]\label{ass:saliency}
The optimizer maintains a diagonal (or Kronecker/low-rank factorized)
statistic $\sal_k \succeq 0$ satisfying
$\norm{\sal_k - \Hess_k}_{\mathrm{op}} \le \delta$
on the relevant neighborhood.
\end{assumption}

For second-moment accumulators, $\sal_k$ is the empirical Fisher diagonal,
a curvature surrogate whose relation to the true Fisher and Gauss--Newton
matrices is well studied \citep{amari1998natural, martens2015kfac}, with
known limitations as a natural-gradient preconditioner
\citep{kunstner2019limitations};
$\delta$ absorbs precisely this approximation gap. The comparability of
trajectory-accumulated saliency to post-hoc Fisher estimates was
established empirically in \citep{moayedikia2026bridging}.
The parameter $\delta$ propagates linearly through all bounds; no result below requires $\delta = 0$. The assumption is nonetheless strong. An operator-norm bound between a diagonal empirical-Fisher statistic and the full Hessian need not have small $\delta$, and $\delta$ is not observable from optimizer byproducts, since it is defined against the true Hessian. Results that carry $\delta$ are informative only to the extent that this gap is small, and where the statistic is rescaled, as in Section~\ref{sec:transformer}, they can be read only as orderings.

\begin{assumption}[Curvature domination by the base statistic]
\label{ass:domination}
There exists $c > 0$ such that
$\nabla^2\loss_k(\theta) \preceq c\,\sal_0$ for every $\theta \in \mathcal{N}$ and every $k \in [K]$, so that in particular $\Hess_k \preceq c\,\sal_0$ at any task solution in $\mathcal{N}$ with a constant independent of which solution training returns (and hence of the penalty coefficient $\mu$),
where $\sal_0$ is the base-model saliency statistic of
\eqref{eq:regularized}.
\end{assumption}

Assumption~\ref{ass:domination} is the only place where the base
statistic $\sal_0$---the object the training-time penalty of
Section~\ref{sec:dynamics} acts on---is related to the task curvatures
$\Hess_k$ that the interference functional \eqref{eq:omega} is measured
in. Without such a relation the two are unconnected quadratic forms, and
a bound on displacement in the $\sal_0$ metric carries no information
about $\Ov_k$; Theorem~\ref{thm:ceiling} is exactly the statement that
this one comparison is sufficient to close the gap. Three points are
worth making about it.

First, it is a Loewner comparison between two curvature objects rather
than a bound on either one, so it is invariant to the overall scaling of
$\sal_0$ (a rescaling $\sal_0 \mapsto a\sal_0$ is absorbed by
$c \mapsto c/a$, and correspondingly by $\mu \mapsto \mu/a$ in
\eqref{eq:regularized}); the normalization of $\sal_0$ to unit operator
norm used in Section~\ref{sec:numerics} therefore costs no generality.

Second, $c$ can be estimated from optimizer byproducts, but only up to the saliency gap $\delta$. By Assumption~\ref{ass:saliency},
$\Hess_k \preceq \sal_k + \delta I$, so it suffices to compare the two
\emph{saliency} statistics: if $\sal_k \preceq c_1 \sal_0$ and
$\sal_0 \succeq \sigma_{\min} I \succ 0$ on the relevant subspace, then
Assumption~\ref{ass:domination} holds with
$c = c_1 + \delta/\sigma_{\min}$. Both $\sal_k$ and $\sal_0$ are
maintained by the optimizer, and for diagonal statistics the comparison
reduces to a coordinatewise ratio. The ratio $c_1$ is therefore computable; the term $\delta/\sigma_{\min}$ is not, because $\delta$ compares against the true Hessian. We accordingly treat $c$ as a problem constant that can be bounded but not read off a training run.

Third, the assumption is substantive, and we record where it can fail:
it requires the base model to carry curvature in every direction in which
a task acquires curvature. A task that develops sensitivity along a
direction that is flat at $\theta_0$ inflates $c$, and in the limit of an
exactly null direction of $\sal_0$ on which $\Hess_k$ is non-degenerate
no finite $c$ exists. For second-moment accumulators, which are strictly
positive in every coordinate that ever receives gradient signal, this
degenerate case does not arise, but $c$ can be large; it enters
Theorem~\ref{thm:ceiling} linearly and is reported rather than assumed
small. The assumption is used only in Theorem~\ref{thm:ceiling}, Proposition~\ref{prop:firstorder}(c), Remark~\ref{rem:isotropic} and
Remark~\ref{rem:ceilingfact}; no result in
Sections~\ref{sec:bound}--\ref{sec:characterization}, and no part of
Theorem~\ref{thm:dynamics}, depends on it.

\begin{definition}[$\varepsilon$-composability]\label{def:composability}
Fix $\varepsilon > 0$ and weights $\bm{\lambda}$. The collection
$\{\tau_k\}_{k \in [K]}$ is \emph{$\varepsilon$-composable at
$\bm{\lambda}$} if
\[
\loss_k\bigl(\merged\bigr) \;\le\; \loss_k(\theta_k) + \varepsilon
\qquad \text{for every } k \in [K].
\]
\end{definition}

The definition is deliberately loss-level rather than function-space
level: it is weaker than the weight disentanglement property of
\citet{ortizjimenez2023task}, which requires the merged model to
\emph{compute the same function} as each expert on its own data support.
Loss-level composability is the property of most direct practical interest, and
Section~\ref{sec:characterization} shows it admits an exact
characterization where function-space disentanglement admits only
sufficient conditions. One caveat bounds what the definition says. It is relative to $\loss_k(\theta_k)$ and is silent on how good $\theta_k$ itself is: the degenerate collection $\tau_k = 0$ is trivially composable. Composability must therefore be read together with the own-task loss, as in the frontier of Section~\ref{sec:numerics}.

\section{The Interference Bound}\label{sec:bound}

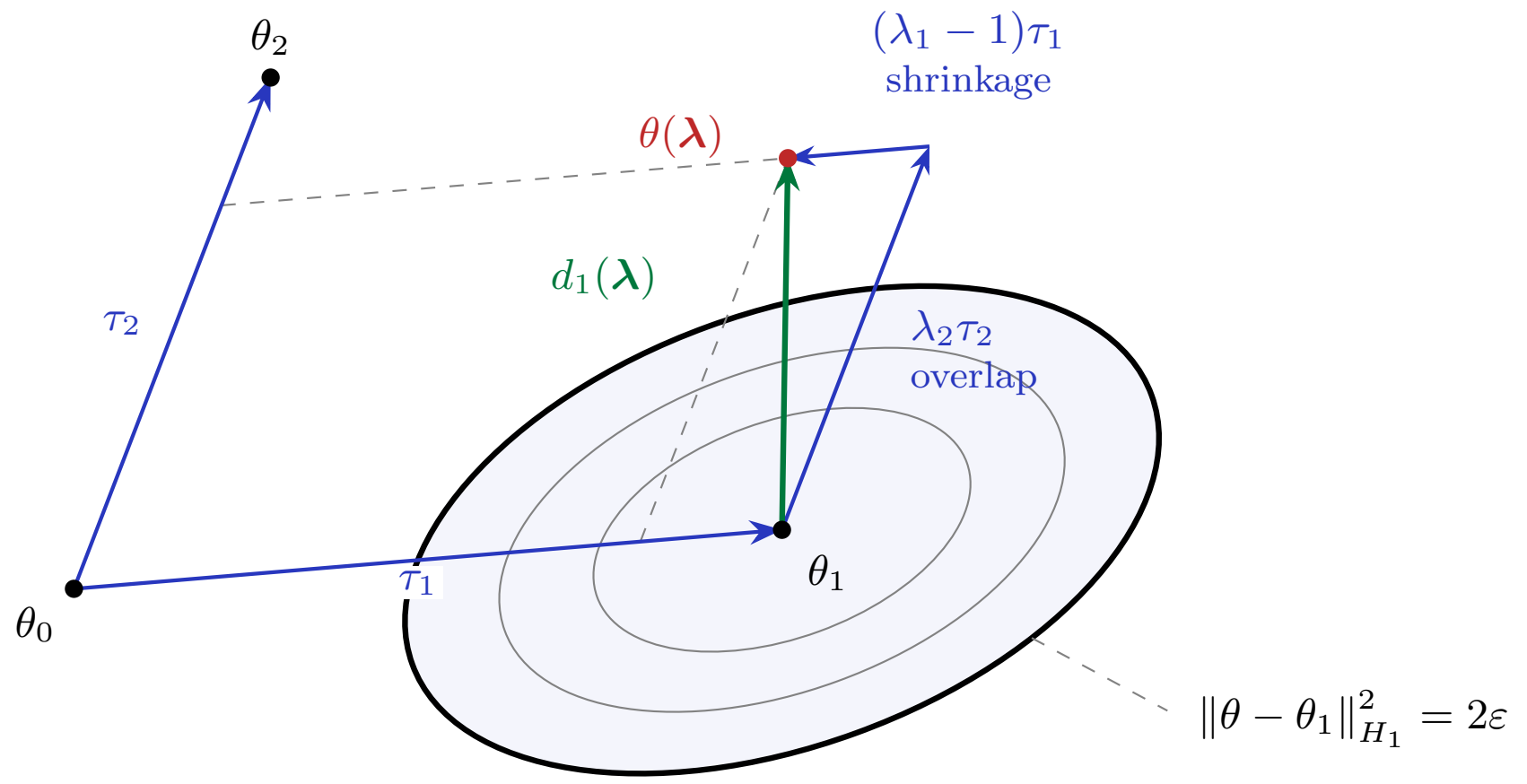
\begin{figure}[!htbp]
\centering
\begin{tikzpicture}[scale=1.45]
\begin{scope}
  \fill[figblue!5, rotate around={20:(3.6,0.3)}] (3.6,0.3) ellipse (2.00 and 1.10);
  \draw[levelset, rotate around={20:(3.6,0.3)}, figgray] (3.6,0.3) ellipse (1.00 and 0.55);
  \draw[levelset, rotate around={20:(3.6,0.3)}, figgray] (3.6,0.3) ellipse (1.50 and 0.82);
  \draw[levelset, rotate around={20:(3.6,0.3)}, very thick] (3.6,0.3) ellipse (2.00 and 1.10);
\end{scope}
\node[lbl, anchor=west] at (5.62,-0.66) {$\norm{\theta-\theta_1}_{\Hess_1}^{2}=2\varepsilon$};
\draw[aux] (4.87,-0.25) -- (5.56,-0.62);

\draw[aux] (0,0) -- (2.88,0.24);
\draw[aux] (2.88,0.24) -- (3.63,2.19);
\draw[aux] (0.75,1.95) -- (3.63,2.19);
\draw[aux] (0,0) -- (0.75,1.95);

\draw[taskvec] (0,0) -- (3.6,0.3);
\node[lblb, fill=white, inner sep=1pt] at (1.75,0.03) {$\tau_1$};
\draw[taskvec] (0,0) -- (1.0,2.6);
\node[lblb, anchor=east] at (0.44,1.35) {$\tau_2$};

\draw[figblue, -{Stealth[length=2mm]}, thick] (3.6,0.3) -- (4.35,2.25);
\draw[figblue, -{Stealth[length=2mm]}, thick] (4.35,2.25) -- (3.63,2.19);
\draw[dispvec] (3.6,0.3) -- (3.63,2.19);

\node[lblb, anchor=west, align=left] at (4.16,1.20) {$\lambda_2\tau_2$\\[-1pt]\scriptsize overlap};
\node[lblb, anchor=south west, align=center] at (3.96,2.40) {$(\lambda_1-1)\tau_1$\\[-1pt]\scriptsize shrinkage};
\node[lblg, anchor=east] at (3.06,1.58) {$d_1(\bm\lambda)$};

\node[pt] at (0,0)      {}; \node[lbl, below left=0pt] at (0,0)      {$\theta_0$};
\node[pt] at (3.6,0.3)  {}; \node[lbl, below right=1pt] at (3.62,0.28) {$\theta_1$};
\node[pt] at (1.0,2.6)  {}; \node[lbl, above=1pt] at (1.0,2.6)       {$\theta_2$};
\node[pt, fill=figred] at (3.63,2.19) {};
\node[lbl, figred, anchor=east] at (3.40,2.30) {$\merged$};
\end{tikzpicture}
\caption{Merge geometry for $K=2$ at $\bm{\lambda}=(0.8,0.75)$: the
displacement $d_1(\bm{\lambda})$ splits into a shrinkage part and an
overlap part whose square in the $\Hess_1$ metric is $\Ov_1$, and the bold
ellipse $\norm{\theta-\theta_1}^2_{\Hess_1}=2\varepsilon$ is the decision
boundary of Theorem~\ref{thm:characterization}.}
\label{fig:geometry}
\end{figure}

Define, for each task $k$, the \emph{interference functional}
\begin{equation}\label{eq:omega}
\Ov_k(\bm{\lambda})
\;=\;
\underbrace{\norm{\sum_{j \ne k} \lambda_j \tau_j}_{\Hess_k}^{2}}_{\text{cross-task curvature overlap}}
\;+\;
\underbrace{2\,(\lambda_k - 1)\, \tau_k^\top \Hess_k \Bigl(\sum_{j \ne k} \lambda_j \tau_j\Bigr)}_{\text{alignment (signed)}}
\;+\;
\underbrace{(\lambda_k - 1)^2 \norm{\tau_k}_{\Hess_k}^{2}}_{\text{shrinkage}} ,
\end{equation}
so that $\Ov_k(\bm{\lambda}) = \norm{d_k(\bm{\lambda})}_{\Hess_k}^2$ by
expansion of \eqref{eq:displacement}.
Figure~\ref{fig:geometry} renders this decomposition for two tasks: the
shrinkage and overlap parts of $d_k$ pull in opposing directions, and it is
their sum, measured in the $\Hess_k$ metric, that the theorems below
control.

\begin{theorem}[Interference bound]\label{thm:bound}
Under Assumptions~\ref{ass:smooth}--\ref{ass:stationary}, for every task
$k$ and all weights $\bm{\lambda} \in \Lambda$,
\[
\loss_k\bigl(\merged\bigr) - \loss_k(\theta_k)
\;\le\;
\tfrac{1}{2}\,\Ov_k(\bm{\lambda})
\;+\; \epsilon_g \norm{d_k(\bm{\lambda})}
\;+\; \tfrac{\rho}{6} \norm{d_k(\bm{\lambda})}^3 .
\]
Consequently, if $\tfrac{1}{2}\Ov_k(\bm{\lambda}) + \epsilon_g
\norm{d_k} + \tfrac{\rho}{6}\norm{d_k}^3 \le \varepsilon$ for all $k$,
the collection is $\varepsilon$-composable at $\bm{\lambda}$.
\end{theorem}

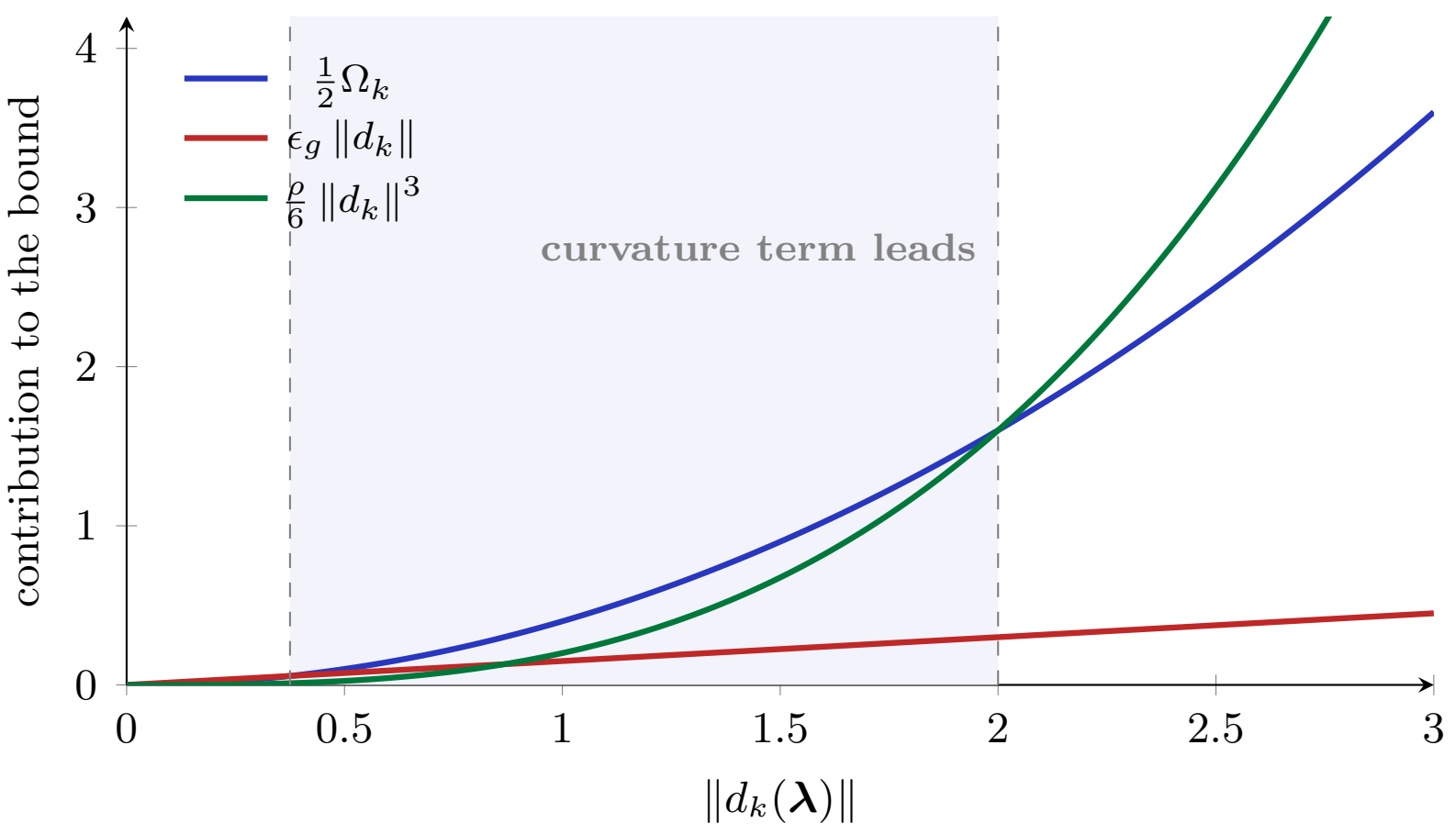
\begin{figure}[!htbp]
\centering
\begin{tikzpicture}
\begin{axis}[
  width=11cm, height=6.4cm,
  xlabel={$\norm{d_k(\bm{\lambda})}$},
  ylabel={contribution to the bound},
  domain=0:3, samples=200,
  xmin=0, xmax=3, ymin=0, ymax=4.2,
  axis lines=left,
  legend style={at={(0.03,0.97)}, anchor=north west, draw=none,
                font=\scriptsize, fill=none},
  tick label style={font=\footnotesize},
  label style={font=\footnotesize},
  every axis plot/.append style={very thick},
]
\addplot[draw=none, fill=figblue!6, forget plot, domain=0.375:2.0] {4.2} \closedcycle;
\addplot[figblue]  {0.4*x^2};   \addlegendentry{$\tfrac12\Ov_k$}
\addplot[figred]   {0.15*x};    \addlegendentry{$\epsilon_g\norm{d_k}$}
\addplot[figgreen] {0.2*x^3};   \addlegendentry{$\tfrac{\rho}{6}\norm{d_k}^3$}
\draw[figgray, dashed] (axis cs:0.375,0) -- (axis cs:0.375,4.2);
\draw[figgray, dashed] (axis cs:2.0,0)   -- (axis cs:2.0,4.2);
\node[font=\scriptsize, figgray, anchor=south, align=center]
      at (axis cs:1.45,2.55) {\textbf{curvature term leads}};
\end{axis}
\end{tikzpicture}
\caption{The three contributions to the bound of Theorem~\ref{thm:bound}
scale as $\norm{d_k}$, $\norm{d_k}^2$ and $\norm{d_k}^3$, so the curvature
term leads only over an intermediate band of displacements (illustrative
constants $\epsilon_g=0.15$, $\rho=1.2$, $\Ov_k=0.8\norm{d_k}^2$; the
crossings move with the constants, the ordering of the exponents does
not).}
\label{fig:boundanatomy}
\end{figure}

\noindent The proof (Appendix~\ref{app:bound}) is a second-order
expansion with integral remainder; for quadratic losses the bound
holds with equality, so it cannot be improved without further
assumptions (Remark~\ref{rem:tightproof}).
The three contributions carry different powers of $\norm{d_k}$, so they
dominate in different regimes (Figure~\ref{fig:boundanatomy}): the
stationarity slack governs very small displacements, the cubic remainder
governs large ones, and the curvature term---the only one the merging
literature has in view---leads over the intermediate band. In the quadratic regime of Section~\ref{sec:characterization} the other two terms vanish identically and the bound is an equality at every displacement, not only in this band.

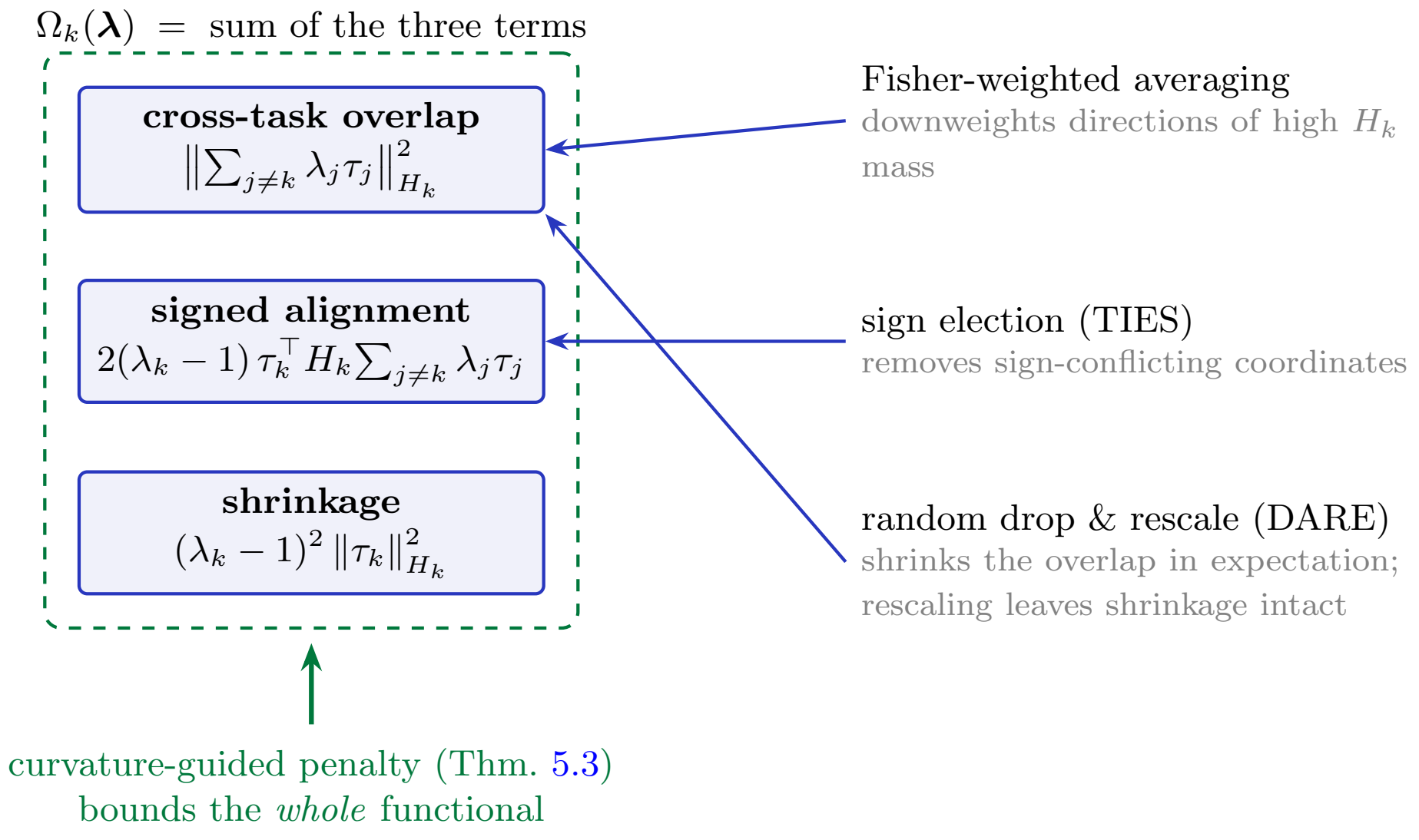
\begin{figure}[!htbp]
\centering
\begin{tikzpicture}[
  term/.style = {draw, thick, rounded corners=2pt, align=center,
                 font=\footnotesize, minimum width=4.0cm, minimum height=1.05cm,
                 fill=figblue!7, draw=figblue},
  op/.style   = {align=left, font=\footnotesize, anchor=west, text width=4.9cm},
  hit/.style  = {figblue, -{Stealth[length=2mm]}, thick},
]
\node[font=\footnotesize] at (0,3.15) {$\Ov_k(\bm{\lambda})\;=\;$ sum of the three terms};

\node[term] (ov) at (0,2.10)
  {\textbf{cross-task overlap}\\[1pt] $\bigl\|\sum_{j\ne k}\lambda_j\tau_j\bigr\|^2_{\Hess_k}$};
\node[term] (al) at (0,0.45)
  {\textbf{signed alignment}\\[1pt] $2(\lambda_k-1)\,\tau_k^\top \Hess_k\!\sum_{j\ne k}\lambda_j\tau_j$};
\node[term] (sh) at (0,-1.20)
  {\textbf{shrinkage}\\[1pt] $(\lambda_k-1)^2\norm{\tau_k}^2_{\Hess_k}$};

\node[op] (fisher) at (4.6,2.35)
  {Fisher-weighted averaging\\[-1pt]
   {\scriptsize\textcolor{figgray}{downweights directions of high $\Hess_k$ mass}}};
\node[op] (ties) at (4.6,0.45)
  {sign election (TIES)\\[-1pt]
   {\scriptsize\textcolor{figgray}{removes sign-conflicting coordinates}}};
\node[op] (dare) at (4.6,-1.45)
  {random drop \& rescale (DARE)\\[-1pt]
   {\scriptsize\textcolor{figgray}{shrinks the overlap in expectation;
    rescaling leaves shrinkage intact}}};

\draw[hit] (fisher.west) -- (ov.east);
\draw[hit] (ties.west)   -- (al.east);
\draw[hit] (dare.west) -- (ov.south east);

\begin{scope}[on background layer]
  \node[draw=figgreen, thick, dashed, rounded corners=4pt,
        fit=(ov)(al)(sh), inner sep=8pt] (all) {};
\end{scope}
\draw[figgreen, -{Stealth[length=2.4mm]}, very thick] (0,-2.85) -- (0,-2.15);
\node[font=\footnotesize, figgreen, align=center, anchor=north] at (0,-2.92)
  {curvature-guided penalty (Thm.~\ref{thm:ceiling})\\ bounds the \emph{whole} functional};
\end{tikzpicture}
\caption{Heuristic reading (Remark~\ref{rem:operators}): each established merging operator can be read as targeting one term of the decomposition~\eqref{eq:omega}, leaving the others uncontrolled, whereas the penalty of Section~\ref{sec:dynamics} bounds the functional as a whole under Assumption~\ref{ass:domination} (schematic; no relative magnitudes are asserted).}
\label{fig:operatormap}
\end{figure}

\begin{remark}[A heuristic reading of existing operators]
\label{rem:operators}
The decomposition \eqref{eq:omega} organizes the merging literature.
Fisher-weighted averaging \citep{matena2022fisher} chooses $\lambda$
coordinate-wise to downweight directions of high $\Hess_k$-mass,
attacking the overlap term with $\sal_k$ substituted for $\Hess_k$.
Sign-election methods \citep{yadav2023ties} zero coordinates on which
task vectors disagree in sign, attacking the conflicting-alignment part of the signed
alignment term. Random dropping \citep{yu2024dare} sparsifies $\tau_j$,
shrinking the overlap term in expectation while rescaling preserves the
shrinkage term. None controls all three terms simultaneously, which is
consistent with the empirically observed task-dependence of their
relative performance: on task collections where the neglected term
dominates, each operator underperforms---the ablation anomalies reported
in \citep{moayedikia2026bridging}, where sign election degrades
performance on RTE while curvature aggregation degrades CoLA, are
consistent with this reading.
Figure~\ref{fig:operatormap} sets the correspondence out term by term.
The correspondence is interpretive and is not a theorem. Fisher-weighted averaging uses coordinate-wise weights and so lies outside the scalar-weight family \eqref{eq:merged}. The signed alignment term carries the factor $(\lambda_k - 1)$: it vanishes at $\lambda_k = 1$ and is active only for averaging-type merges with $\lambda_k < 1$, where it is positive exactly when $\tau_k$ conflicts with the sibling sum in the $\Hess_k$ inner product. In particular, for $K = 2$ and $\bm{\lambda} = (1,1)$ one has $\Ov_k = \norm{\tau_j}_{\Hess_k}^2$ with $j \ne k$, which does not depend on the alignment between $\tau_k$ and $\tau_j$ at all; for $K \ge 3$ the cross terms among siblings sit inside the overlap term.
\end{remark}

\begin{remark}[Computability]\label{rem:computable}
Every quantity in $\Ov_k$ except $\Hess_k$ is known at merge time, and
Assumption~\ref{ass:saliency} licenses replacing $\Hess_k$ by the
trajectory statistic $\sal_k$ at additive cost
$\delta \sum_k \norm{d_k}^2$: the functional is computable from
optimizer byproducts, without data access or a separate estimation pass.
This is what makes the criterion usable as a pre-merge screen; because $\delta$ is itself unobserved, the reliability of the screen rests on Assumption~\ref{ass:saliency}.
\end{remark}

\begin{algorithm}[!htbp]
\caption{Pre-merge composability screening: evaluates the certificate of
Theorem~\ref{thm:bound} from stored optimizer statistics.}
\label{alg:screen}
\begin{algorithmic}[1]
\Require base model $\theta_0$; fine-tuned models $\{\theta_k\}_{k\in[K]}$;
  saliency statistics in factored form $\sal_k^{(r)}=A_kA_k^\top$ with
  $A_k\in\R^{d\times r}$; candidate weights $\bm{\lambda}\in\Lambda$;
  tolerance $\varepsilon$; constants $\delta,\epsilon_g,\rho$ and
  truncation errors $\epsilon_r(\sal_k)$
\Ensure certificate $\kappa(\bm{\lambda})$ and a verdict
\State $\tau_k \gets \theta_k-\theta_0$ for every $k$ \algcomment{task vectors}
\State $\merged \gets \theta_0+\sum_{k}\lambda_k\tau_k$
       \algcomment{Eq.~\eqref{eq:merged}; formed, not evaluated}
\For{$k=1,\dots,K$}
  \State $d_k \gets \merged-\theta_k$ \algcomment{Eq.~\eqref{eq:displacement}}
  \State $z_k \gets A_k^\top d_k$ \algcomment{$O(dr)$: the only large product}
  \State $\widehat{\Ov}_k \gets \norm{z_k}^2$
         \algcomment{$=\norm{d_k}^2_{\sal_k^{(r)}}$, Cor.~\ref{cor:factorized}(a)}
  \State $\kappa_k \gets \tfrac12\widehat{\Ov}_k
           +\tfrac12\bigl(\delta+\epsilon_r(\sal_k)\bigr)\norm{d_k}^2
           +\epsilon_g\norm{d_k}+\tfrac{\rho}{6}\norm{d_k}^3$
         \algcomment{Thm.~\ref{thm:bound} and Cor.~\ref{cor:factorized}(a)}
\EndFor
\State $\kappa(\bm{\lambda})\gets\max_{k\in[K]}\kappa_k$
\If{$\kappa(\bm{\lambda})\le\varepsilon$}
  \State \textbf{return} $\bigl(\kappa(\bm{\lambda}),\ \text{\emph{$\varepsilon$-composable}}\bigr)$
\Else
  \State \textbf{return} $\bigl(\kappa(\bm{\lambda}),\ \text{\emph{not certified}}\bigr)$
\EndIf
\end{algorithmic}
\end{algorithm}

Algorithm~\ref{alg:screen} states the resulting procedure. It costs
$O(Kdr)$ time and $O(dr)$ memory beyond the models themselves, reads no
data from any $\mathcal{D}_k$, and runs no forward or backward pass; when
$\sal_k$ is diagonal rather than factored, line~5 becomes a Hadamard
product and the cost drops to $O(Kd)$. Only $\delta$, $\epsilon_g$ and
$\rho$ are not exactly computable, and they enter the slack alone, so
setting them to zero returns the leading-order screen
$\max_k\tfrac12\widehat{\Ov}_k\le\varepsilon$ used in
Section~\ref{sec:numerics}. In the quadratic regime of
Section~\ref{sec:characterization} a negative verdict is a refusal rather
than a failure to certify, by Theorem~\ref{thm:characterization}.

\section{Characterization in the \texorpdfstring{Quadratic}{Quadratic} Regime}
\label{sec:characterization}

We now restrict to quadratic task losses,
$\loss_k(\theta) = \loss_k(\theta_k) + \tfrac12
\norm{\theta - \theta_k}_{\Hess_k}^2$, the regime that arises exactly for
NTK-linearized fine-tuning under a squared loss \citep{jacot2018ntk, lee2019wide,
ortizjimenez2023task} (for other losses, cross-entropy included, linearizing the model gives a convex but non-quadratic objective, of which the quadratic form is the local second-order model) and approximately whenever fine-tuning remains in a
locally quadratic basin; that fine-tunes from a shared initialization tend
to occupy a common basin is the content of the linear-mode-connectivity
literature \citep{frankle2020linear, entezari2022role}. In this
regime we take each $\theta_k$ to be an exact local minimizer of
$\loss_k$, so that $\epsilon_g = 0$ in
Assumption~\ref{ass:stationary} and $\Hess_k \succeq 0$; the
characterization below in fact requires only symmetry of $\Hess_k$,
positive semi-definiteness playing no role in the equivalence.

\begin{theorem}[Characterization]\label{thm:characterization}
For quadratic task losses, the collection $\{\tau_k\}$ is
$\varepsilon$-composable at $\bm{\lambda}$ if and only if
\[
\max_{k \in [K]} \; \Ov_k(\bm{\lambda}) \;\le\; 2\varepsilon .
\]
\end{theorem}

\noindent See Appendix~\ref{app:characterization} for the proof.

\begin{proposition}[Convexity of the criterion in $\bm{\lambda}$]
\label{prop:convex}
For any symmetric positive semi-definite $\{\sal_k\}$, the map
$\bm{\lambda}\mapsto\max_{k\in[K]}\norm{d_k(\bm{\lambda})}^2_{\sal_k}$ is
convex on $\R^K$.
\end{proposition}

\begin{proof}
By \eqref{eq:displacement}, $d_k(\bm{\lambda})$ is affine in
$\bm{\lambda}$, so $\bm{\lambda}\mapsto\norm{d_k(\bm{\lambda})}^2_{\sal_k}$
is the composition of an affine map with the convex quadratic
$x\mapsto x^\top \sal_k x$ and is therefore convex; a pointwise maximum of
finitely many convex functions is convex.
\end{proof}

\noindent Proposition~\ref{prop:convex} turns
Theorem~\ref{thm:characterization} from a test into a design objective:
over a convex admissible set $\Lambda$, the weights that minimize the
criterion can be found by projected subgradient descent
(Algorithm~\ref{alg:weights}), at the standard $O(1/\sqrt{S})$ rate for
nonsmooth convex minimization.

\begin{algorithm}[!htbp]
\caption{Selecting merging weights by minimizing the criterion of
Theorem~\ref{thm:characterization} over $\Lambda$.}
\label{alg:weights}
\begin{algorithmic}[1]
\Require $\theta_0$, $\{\theta_k\}$, factors $\{A_k\}$, convex admissible
  set $\Lambda\subset\R^K_{\ge0}$, step sizes $\{\alpha_s\}$, iterations $S$
\Ensure weights approximately minimizing
  $\max_k\widehat{\Ov}_k(\bm{\lambda})$
\State $\bm{\lambda}^{(0)}\gets(1,\dots,1)$
\For{$s=0,\dots,S-1$}
  \State $k_s\gets\arg\max_{k}\widehat{\Ov}_k(\bm{\lambda}^{(s)})$
         \algcomment{active task; Alg.~\ref{alg:screen}, lines~4--6}
  \State $d\gets\theta(\bm{\lambda}^{(s)})-\theta_{k_s}$
  \State $g_j \gets 2\,\tau_j^\top A_{k_s}A_{k_s}^\top d$ for $j\in[K]$
         \algcomment{subgradient, $O(Kdr)$}
  \State $\bm{\lambda}^{(s+1)}\gets\Pi_\Lambda\bigl(\bm{\lambda}^{(s)}-\alpha_s g\bigr)$
\EndFor
\State \textbf{return} $\bm{\lambda}^{(S)}$
\end{algorithmic}
\end{algorithm}

\begin{example}[Tightness and a phase boundary]\label{ex:tight}
Let $d = 2$, $K = 2$, $\Hess_1 = \Hess_2 = \mathrm{diag}(h, h')$ with
$h, h' > 0$, and $\tau_1 = (a, 0)$, $\tau_2 = (b\cos\phi,
b\sin\phi)$. At $\lambda_1 = \lambda_2 = 1$, $\Ov_1 = b^2(h\cos^2\phi +
h'\sin^2\phi)$: composability is governed by the angle $\phi$ between
task vectors \emph{as measured in the curvature metric}, not the
Euclidean angle. The construction exhibits, for any $\varepsilon$, task
pairs that are Euclidean-orthogonal yet fail $\varepsilon$-composability
($\phi = \pi/2$ with $h'$ large) and task pairs with substantial
Euclidean overlap that compose freely ($\phi = 0$ with $h$ small);
Figure~\ref{fig:metric} draws the two regimes side by side.
Which angle is the more favourable depends on the anisotropy: with $\norm{\tau_2} = b$ fixed, $\Ov_1$ is smaller at $\phi = \pi/2$ when $h > h'$ and smaller at $\phi = 0$ when $h < h'$.
The computation is given in Appendix~\ref{app:example}.
\end{example}

\begin{figure}[!htbp]
\centering
\begin{tikzpicture}[scale=1.22]
\begin{scope}
  \fill[figblue!5] (2.4,0) ellipse (2.50 and 1.20);
  \draw[levelset, figgray] (2.4,0) ellipse (1.25 and 0.60);
  \draw[levelset, figgray] (2.4,0) ellipse (1.88 and 0.90);
  \draw[levelset, very thick] (2.4,0) ellipse (2.50 and 1.20);
  \draw[figgray, dotted, thin] (2.4,-1.30) -- (2.4,1.15);
  \draw[figgray, dotted, thin] (-0.05,0) -- (4.85,0);
  \node[lbl, figgray, anchor=north] at (2.4,-1.32) {stiff};
  \node[lbl, figgray, anchor=west] at (4.95,0) {soft};
  \draw[taskvec] (0,0) -- (2.4,0);
  \node[lblb, fill=white, inner sep=1pt] at (1.25,0) {$\tau_1$};
  \draw[taskvec] (0,0) -- (0,1.8);
  \node[lblb, anchor=east] at (-0.08,0.95) {$\tau_2$};
  \draw[aux] (0,1.8) -- (2.4,1.8);
  \draw[aux] (2.4,0) -- (2.4,1.8);
  \draw[dispvec] (2.4,0) -- (2.4,1.8);
  \node[lblg, anchor=west] at (2.48,0.95) {$d_1$};
  \node[pt] at (0,0)   {}; \node[lbl, below left=0pt] at (0,0) {$\theta_0$};
  \node[pt] at (2.4,0) {}; \node[lbl, fill=white, inner sep=1pt, anchor=north east] at (2.34,-0.06) {$\theta_1$};
  \node[pt] at (0,1.8) {}; \node[lbl, above left=-1pt] at (0,1.8) {$\theta_2$};
  \node[pt, fill=figred] at (2.4,1.8) {};
  \node[lbl, figred, anchor=south] at (2.4,1.88) {$\merged$};
  \node[lbl, align=center] at (2.4,-2.10)
       {(a) $\angle(\tau_1,\tau_2)=90^\circ$\\ \textbf{not} $\varepsilon$-composable};
\end{scope}
\begin{scope}[xshift=7.0cm]
  \fill[figblue!5] (2.4,0) ellipse (2.50 and 1.20);
  \draw[levelset, figgray] (2.4,0) ellipse (1.25 and 0.60);
  \draw[levelset, figgray] (2.4,0) ellipse (1.88 and 0.90);
  \draw[levelset, very thick] (2.4,0) ellipse (2.50 and 1.20);
  \draw[figgray, dotted, thin] (2.4,-1.30) -- (2.4,1.15);
  \draw[figgray, dotted, thin] (-0.05,0) -- (4.85,0);
  \node[lbl, figgray, anchor=north] at (2.4,-1.32) {stiff};
  \node[lbl, figgray, anchor=west] at (4.95,0) {soft};
  \draw[taskvec] (0,0) -- (2.4,0);
  \node[lblb, fill=white, inner sep=1pt] at (1.15,-0.02) {$\tau_1$};
  \draw[taskvec] (0,0) -- (1.739,0.466);
  \node[lblb, anchor=south east] at (0.95,0.29) {$\tau_2$};
  \draw[aux] (1.739,0.466) -- (4.139,0.466);
  \draw[aux] (2.4,0) -- (4.139,0.466);
  \draw[dispvec] (2.4,0) -- (4.139,0.466);
  \node[lblg, anchor=north west] at (3.25,0.20) {$d_1$};
  \node[pt] at (0,0) {}; \node[lbl, below left=0pt] at (0,0) {$\theta_0$};
  \node[pt] at (2.4,0) {}; \node[lbl, fill=white, inner sep=1pt, anchor=north east] at (2.34,-0.08) {$\theta_1$};
  \node[pt] at (1.739,0.466) {}; \node[lbl, anchor=south east, inner sep=2pt] at (1.72,0.52) {$\theta_2$};
  \node[pt, fill=figgreen] at (4.139,0.466) {};
  \node[lblg, anchor=south west] at (4.22,0.60) {$\merged$};
  \node[lbl, align=center] at (2.4,-2.10)
       {(b) $\angle(\tau_1,\tau_2)=15^\circ$\\ $\varepsilon$-composable};
\end{scope}
\end{tikzpicture}
\caption{Drawn for $h < h'$ (soft along $\tau_1$, stiff across it). With curvature, tolerance, $\norm{\tau_2}$ and
$\bm{\lambda}=(1,1)$ all held fixed, Euclidean-orthogonal task vectors~(a)
can fail $\varepsilon$-composability while strongly overlapping ones~(b)
satisfy it, because the criterion measures the angle in the $\Hess_k$
metric rather than the Euclidean one.}
\label{fig:metric}
\end{figure}

\begin{remark}
Theorem~\ref{thm:characterization} upgrades the loss-landscape bounds of
\citet{lu2026geometry} and the second-order analyses surveyed in
\citep{yang2026model} from sufficient conditions to a criterion that is
also necessary, at the price of the quadratic restriction. The upgrade is immediate---for quadratic losses the functional \emph{is} the excess loss---and its value is as a reference point rather than as a technical advance. The example
shows the criterion is genuinely metric: approximate orthogonality of
task vectors, the condition emphasized in prior analyses, is neither
necessary nor sufficient once curvature is anisotropic.
\end{remark}

\section{Merge-Aware Training Dynamics}\label{sec:dynamics}

The results above take $\{\tau_k\}$ as given. We now modify how they are
produced. Consider fine-tuning task $k$ under the regularized objective
\begin{equation}\label{eq:regularized}
J_k(\theta) \;=\; \loss_k(\theta) \;+\;
\frac{\mu}{2}\,(\theta - \theta_0)^\top \sal_0\,(\theta - \theta_0),
\end{equation}
where $\sal_0$ is the saliency statistic of the \emph{base} model and
$\mu > 0$. Throughout this section $\sal_0$ is symmetric positive
semi-definite with finite operator norm, which holds by construction
for accumulated second-moment statistics; this is all that parts
(i) and (iii) of Theorem~\ref{thm:dynamics} use, the approximation
guarantee of Assumption~\ref{ass:saliency} entering only where
$\sal_0$ stands in for curvature, as in
Corollary~\ref{cor:factorized}. The penalty discourages
movement along directions of high base-model curvature. The intent is
that these are the directions along which any \emph{other} task's loss
is most sensitive, so that constraining them suppresses $\Ov_j$ for
every prospective sibling task $j$ simultaneously, without access to the
siblings; Assumption~\ref{ass:domination} is the hypothesis under which
that intent is correct, and Theorem~\ref{thm:ceiling} is its
quantitative form. We are careful to keep the two separate: parts
(i)--(iii) of Theorem~\ref{thm:dynamics} concern the base metric
$\sal_0$ alone and are unconditional in this respect, while the passage
to $\Ov_j$ is isolated in Theorem~\ref{thm:ceiling} and carries the
constant $c$.
Algorithm~\ref{alg:mat} gives the resulting procedure. It is
optimizer-agnostic---lines~4--5 modify the gradient, not the update rule,
so any adaptive optimizer may replace the descent step of line~6 provided
$\sal_0$ remains available---and it keeps the penalty gradient
deterministic, which is what allows the task-loss oracle alone to carry
the stochasticity in Theorem~\ref{thm:dynamics}(ii). The statistic
$\sal_0$ is taken as given here; obtaining it as a training byproduct is
the subject of \citep{moayedikia2026bridging}, and none of the analysis
below depends on which accumulation scheme produces it. In the transformer study of Section~\ref{sec:transformer} it is not a training byproduct: it is estimated there by a separate diagonal-Fisher pass over generic text.

\begin{algorithm}[!htbp]
\caption{Merge-aware fine-tuning of task $k$ under the regularized
objective~\eqref{eq:regularized}.}
\label{alg:mat}
\begin{algorithmic}[1]
\Require base model $\theta_0$; base saliency in factored form
  $\sal_0^{(r)}=A_0A_0^\top$ with $A_0\in\R^{d\times r}$; penalty
  coefficient $\mu>0$; horizon $T$; step size $\gamma_t=\gamma/\sqrt{T}$
\Ensure $\hat\theta_k$ and its interference budget certificate
\State $\theta^{(0)}\gets\theta_0$
       \algcomment{initialization at $\theta_0$ is required by Thm.~\ref{thm:dynamics}(ii)}
\For{$t=0,\dots,T-1$}
  \State sample $\xi_t\sim\mathcal{D}_k$;\quad
         $g_t\gets\nabla_\theta\,\ell(\theta^{(t)},\xi_t)$
         \algcomment{stochastic task gradient}
  \State $u_t\gets A_0^\top(\theta^{(t)}-\theta_0)$
  \State $p_t\gets \mu\,A_0u_t$ \algcomment{exact penalty gradient, $O(dr)$}
  \State $\theta^{(t+1)}\gets\theta^{(t)}-\gamma_t\,(g_t+p_t)$
  \If{monitoring}
    \State $\Phi_t\gets\norm{u_t}^2$;\quad
           $B_t\gets\tfrac{2}{\mu}\bigl(\loss_k(\theta_0)-\loss_k(\theta^{(t)})\bigr)$
           \algcomment{Thm.~\ref{thm:dynamics}(ii): $\Phi_t\le B_t$ (in expectation under SGD)}
  \EndIf
\EndFor
\State $\hat\theta_k\gets\theta^{(T)}$
\State \textbf{return} $\hat\theta_k$ and
       $\tfrac{2}{\mu}\bigl(\loss_k(\theta_0)-\loss_k(\hat\theta_k)\bigr)$
       \algcomment{Thm.~\ref{thm:dynamics}(iii); stationarity not required}
\end{algorithmic}
\end{algorithm}

\begin{theorem}[Convergence and anytime interference budget]\label{thm:dynamics}
Let Assumptions~\ref{ass:smooth}--\ref{ass:saliency} hold, and write
$L_\mu = L + \mu \norm{\sal_0}_{\mathrm{op}}$.
\begin{itemize}
\item[(i)] Suppose in addition that $\loss_k$ is bounded below by
$\loss_k^\star$ and that Assumption~\ref{ass:smooth} holds on a set
containing the iterates (e.g.\ $\mathcal{N} = \R^d$). Let
$\{\theta^{(t)}\}_{t \ge 0}$ be produced by stochastic gradient
descent on $J_k$ initialized at $\theta^{(0)} = \theta_0$, with
gradient oracle $g_t$ satisfying
$\E[g_t \mid \mathcal{F}_{t-1}] = \nabla J_k(\theta^{(t)})$ and
$\E[\norm{g_t - \nabla J_k(\theta^{(t)})}^2 \mid \mathcal{F}_{t-1}]
\le \sigma^2$, and constant step size $\gamma_t = \gamma/\sqrt{T}$
with $\gamma > 0$. Then for every $T \ge \gamma^2 L_\mu^2$,
\[
\min_{0 \le t < T} \; \E\, \norm{\nabla J_k(\theta^{(t)})}^2
\;\le\;
\frac{1}{\sqrt{T}}
\left(
\frac{2\bigl(\loss_k(\theta_0) - \loss_k^\star\bigr)}{\gamma}
+ L_\mu\, \gamma\, \sigma^2
\right)
\;=\; O\!\left(\frac{1 + \mu\norm{\sal_0}_{\mathrm{op}}}{\sqrt{T}}\right).
\]
\item[(ii)] \emph{(Anytime interference budget.)} Consider the
surrogate interference
$\Phi(\theta) = \norm{\theta - \theta_0}_{\sal_0}^2$. Along the
gradient flow of $J_k$, or along gradient descent with step size
$\gamma \le 1/L_\mu$, both initialized at $\theta_0$, the iterate at
every time $t \ge 0$ satisfies
\[
\Phi(\theta^{(t)})
\;\le\;
\frac{2}{\mu}\bigl(\loss_k(\theta_0) - \loss_k(\theta^{(t)})\bigr)
\;\le\;
\frac{2}{\mu}\bigl(\loss_k(\theta_0) - \loss_k^\star\bigr).
\]
Under the stochastic dynamics of part (i) with $T \ge \gamma^2 L_\mu^2$,
the same holds in expectation up to an additive inflation that does not depend on $T$ and vanishes as $\gamma \to 0$: for all
$t \le T$,
\[
\E\, \Phi(\theta^{(t)})
\;\le\;
\frac{2}{\mu}\bigl(\loss_k(\theta_0) - \E\,\loss_k(\theta^{(t)})\bigr)
+ \frac{\gamma^2 L_\mu \sigma^2}{\mu}.
\]
\item[(iii)] any point $\hat\theta_k$ satisfying
$J_k(\hat\theta_k) \le J_k(\theta_0)$---in particular any limit point
of the gradient flow or deterministic gradient descent of part
(ii)---satisfies
$\norm{\hat\theta_k - \theta_0}_{\sal_0}^2 \le
\tfrac{2}{\mu}\bigl(\loss_k(\theta_0) - \loss_k(\hat\theta_k)\bigr)$,
so the interference budget of the resulting task vector is explicitly
controlled by $\mu$. Stationarity is not required.
\end{itemize}
Parts (i), (ii), and (iii) are proven in
Appendices~\ref{app:sgd} and~\ref{app:budget}.
\end{theorem}

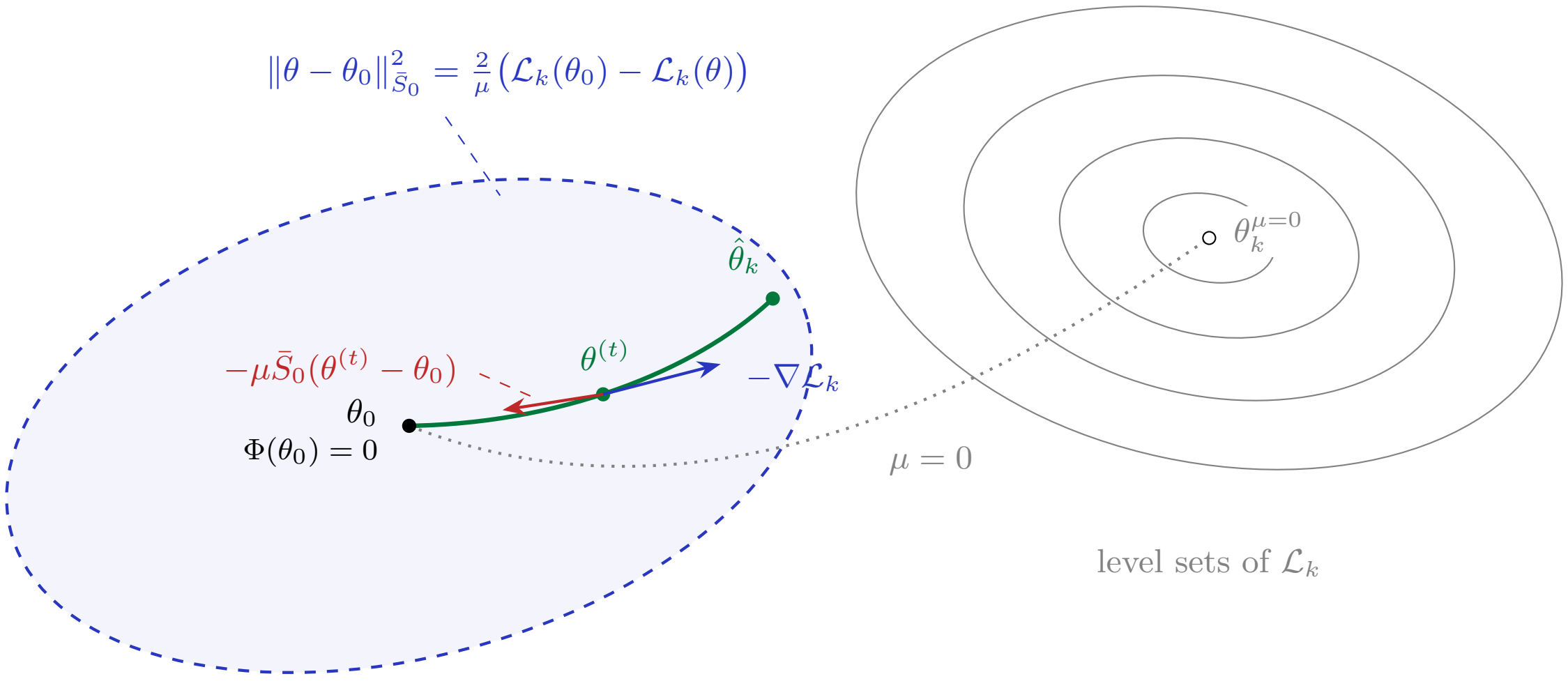
\begin{figure}[!htbp]
\centering
\begin{tikzpicture}[scale=1.15]
\begin{scope}[rotate around={-12:(6.6,1.55)}]
  \draw[levelset, figgray] (6.6,1.55) ellipse (0.55 and 0.36);
  \draw[levelset, figgray] (6.6,1.55) ellipse (1.25 and 0.80);
  \draw[levelset, figgray] (6.6,1.55) ellipse (2.05 and 1.30);
  \draw[levelset, figgray] (6.6,1.55) ellipse (2.95 and 1.85);
\end{scope}
\node[lbl, figgray, anchor=north] at (6.60,-0.90) {level sets of $\loss_k$};

\fill[figblue!5, rotate around={15:(0,0)}] (0,0) ellipse (3.40 and 1.90);
\draw[budget, rotate around={15:(0,0)}] (0,0) ellipse (3.40 and 1.90);
\node[lblb, align=left, anchor=south west] at (-1.30,2.60)
  {$\norm{\theta-\theta_0}_{\sal_0}^{2}
    =\tfrac{2}{\mu}\bigl(\loss_k(\theta_0)-\loss_k(\theta)\bigr)$};
\draw[aux, figblue] (0.30,2.55) -- (0.75,1.90);

\draw[figgray, thick, dotted] (0,0) .. controls (2.4,-0.95) and (4.8,0.30) .. (6.6,1.55);
\node[ptopen] at (6.6,1.55) {};
\node[lbl, figgray, fill=white, inner sep=1.5pt, anchor=west] at (6.75,1.60) {$\theta_k^{\mu=0}$};
\node[lbl, figgray, fill=white, inner sep=1.5pt, anchor=north east] at (4.70,-0.12) {$\mu=0$};

\draw[traj] (0,0) .. controls (1.30,0.02) and (2.40,0.48) .. (3.00,1.05);
\node[pt, fill=figgreen] at (3.00,1.05) {};
\node[lblg, anchor=south east] at (3.02,1.14) {$\hat\theta_k$};
\node[pt] at (0,0) {};
\node[lbl, anchor=east, align=right] at (-0.14,-0.06)
  {$\theta_0$\\[-1pt]\scriptsize$\Phi(\theta_0)=0$};

\node[pt, fill=figgreen] at (1.60,0.26) {};
\node[lblg, anchor=south] at (1.62,0.34) {$\theta^{(t)}$};
\draw[figblue, -{Stealth[length=2.2mm]}, thick] (1.60,0.26) -- (2.57,0.51);
\draw[figred,  -{Stealth[length=2.2mm]}, thick] (1.60,0.26) -- (0.77,0.13);
\node[lblb, anchor=west] at (2.66,0.40) {$-\nabla\loss_k$};
\node[lbl, figred, anchor=east] at (0.52,0.48) {$-\mu\sal_0(\theta^{(t)}-\theta_0)$};
\draw[aux, figred] (0.58,0.43) -- (1.06,0.22);
\end{tikzpicture}
\caption{Along the regularized dynamics the task gradient drives the
iterate outward while the penalty pulls it back in the $\sal_0$ metric, so
the trajectory stays inside the budget of
Theorem~\ref{thm:dynamics}(ii) at every $t$ and equilibrates where the two
forces cancel, whereas the $\mu=0$ run (dotted) leaves the region.}
\label{fig:budget-schematic}
\end{figure}

\begin{remark}[Why the budget is a ceiling, not a decrease]
\label{rem:flowidentity}
Part (ii) bounds $\Phi$ but does not claim it decreases, and it
cannot: since training initializes at $\theta_0$, where
$\Phi(\theta_0) = 0$ is the global minimum of $\Phi$, the surrogate
interference starts at zero, cannot fall below it, and need not be monotone along the trajectory. The dynamics
are transparent along the gradient flow, where a direct computation
gives the \emph{identity}
\[
\frac{d}{dt}\Phi(\theta)
= -2\mu \norm{\sal_0(\theta - \theta_0)}^2
- 2(\theta - \theta_0)^\top \sal_0 \nabla\loss_k(\theta),
\]
a restoring pull from the penalty against the outward drive of the
task gradient. At any stationary point of $J_k$ the two cancel
($\mu\sal_0(\theta - \theta_0) = -\nabla\loss_k(\theta)$), so
$\tfrac{d}{dt}\Phi = 0$: the trajectory comes to rest inside the budget of part (ii), strictly inside whenever $J_k$ has decreased, since the slack in the budget inequality is $\tfrac{2}{\mu}\bigl(J_k(\theta_0) - J_k(\theta)\bigr)$, and $\Phi$ is strictly
decreasing in particular wherever the restoring pull dominates,
$\mu\norm{\sal_0(\theta - \theta_0)} > \norm{\nabla\loss_k(\theta)}$. The value of part (ii) is thus
not monotonicity but the uniform-in-$t$ ceiling: at no point of
training does the accumulated interference exceed what $\mu$ permits.
Figure~\ref{fig:budget-schematic} shows the two forces and the region
they confine the trajectory to.
\end{remark}

Theorem~\ref{thm:dynamics} controls the surrogate $\Phi$, which is
displacement from $\theta_0$ measured in the base metric $\sal_0$. The
quantity the merging bound of Section~\ref{sec:bound} is stated in is
$\Ov_k$, measured in the task curvature $\Hess_k$ and assembled from the
task vectors of \emph{all} tasks in the collection. The next theorem
closes that gap: under Assumption~\ref{ass:domination}, running
Algorithm~\ref{alg:mat} on each task separately converts the per-task
budgets of Theorem~\ref{thm:dynamics}(iii) into an explicit ceiling on
the interference functional of the merged collection, scaling as
$1/\mu$.

\begin{theorem}[Merge-time interference ceiling]\label{thm:ceiling}
Let Assumptions~\ref{ass:smooth}--\ref{ass:domination} hold. Suppose
every task $j \in [K]$ is fine-tuned from $\theta_0$ under the
objective \eqref{eq:regularized} with the same base statistic
$\sal_0$ and the same penalty coefficient $\mu > 0$, returning
$\hat\theta_j = \theta_0 + \tau_j$ with
$J_j(\hat\theta_j) \le J_j(\theta_0)$, and write
\[
\Delta_j \;=\; \loss_j(\theta_0) - \loss_j(\hat\theta_j) \;\ge\; 0
\]
for the task-loss descent achieved on task $j$. Then for every
$k \in [K]$ and every $\bm{\lambda} \in \Lambda$,
\begin{equation}\label{eq:ceiling}
\Ov_k(\bm{\lambda})
\;\le\;
\frac{2c}{\mu}
\Bigl(
|\lambda_k - 1|\sqrt{\Delta_k}
\;+\; \sum_{j \ne k} \lambda_j \sqrt{\Delta_j}
\Bigr)^{2} .
\end{equation}
In particular, writing $\Delta = \max_{j \in [K]} \Delta_j$ and
$e_k$ for the $k$-th standard basis vector,
\begin{equation}\label{eq:ceiling-simple}
\max_{k \in [K]} \Ov_k(\bm{\lambda})
\;\le\;
\frac{2c\,\Delta}{\mu}\,
\max_{k \in [K]} \norm{\bm{\lambda} - e_k}_1^2 ,
\end{equation}
and, combining with Theorem~\ref{thm:bound}, the collection is
$\varepsilon$-composable at $\bm{\lambda}$ whenever
\[
\frac{c}{\mu}
\Bigl(|\lambda_k - 1|\sqrt{\Delta_k}
+ \sum_{j \ne k} \lambda_j \sqrt{\Delta_j}\Bigr)^{2}
+ \epsilon_g \norm{d_k(\bm{\lambda})}
+ \frac{\rho}{6}\norm{d_k(\bm{\lambda})}^3
\;\le\; \varepsilon
\qquad \text{for every } k \in [K].
\]
Here $\epsilon_g$ is the stationarity slack of $\hat\theta_k$ with respect to the task loss $\loss_k$, which regularized training does not make small; Proposition~\ref{prop:firstorder} replaces that term by its exact value.
\end{theorem}

\noindent The proof is in Appendix~\ref{app:ceiling}.

\begin{proposition}[First-order term under regularized training]
\label{prop:firstorder}
In the setting of Theorem~\ref{thm:ceiling}, write
$B_k(\bm{\lambda}) = |\lambda_k - 1|\sqrt{\Delta_k}
+ \sum_{j \ne k} \lambda_j \sqrt{\Delta_j}$
for the bracket of \eqref{eq:ceiling}. Then for every $k \in [K]$ and
every $\bm{\lambda} \in \Lambda$:
\begin{itemize}
\item[(a)] $\nabla\loss_k(\hat\theta_k)
= \nabla J_k(\hat\theta_k) - \mu\,\sal_0\tau_k$, and consequently
\[
\loss_k\bigl(\merged\bigr) - \loss_k(\hat\theta_k)
\;\le\;
\norm{\nabla J_k(\hat\theta_k)}\,\norm{d_k(\bm{\lambda})}
\;-\; \mu\,\tau_k^\top \sal_0\, d_k(\bm{\lambda})
\;+\; \tfrac12\,\Ov_k(\bm{\lambda})
\;+\; \tfrac{\rho}{6}\norm{d_k(\bm{\lambda})}^3 .
\]
\item[(b)] The signed first-order term satisfies
$\bigl|\mu\,\tau_k^\top \sal_0\, d_k(\bm{\lambda})\bigr|
\le 2\sqrt{\Delta_k}\,B_k(\bm{\lambda})$.
\item[(c)] The collection is $\varepsilon$-composable at $\bm{\lambda}$
whenever, for every $k \in [K]$,
\[
\norm{\nabla J_k(\hat\theta_k)}\,\norm{d_k(\bm{\lambda})}
+ 2\sqrt{\Delta_k}\,B_k(\bm{\lambda})
+ \frac{c}{\mu}\,B_k(\bm{\lambda})^2
+ \frac{\rho}{6}\norm{d_k(\bm{\lambda})}^3
\;\le\; \varepsilon .
\]
\end{itemize}
\end{proposition}

\begin{proof}
(a) Differentiating \eqref{eq:regularized} gives
$\nabla J_k(\theta) = \nabla\loss_k(\theta) + \mu\,\sal_0(\theta - \theta_0)$;
evaluate at $\hat\theta_k = \theta_0 + \tau_k$. The expansion
\eqref{eq:taylor} at $\theta_k = \hat\theta_k$, with the remainder bound
established in the proof of Theorem~\ref{thm:bound}, reads
$\loss_k(\merged) - \loss_k(\hat\theta_k)
\le \nabla\loss_k(\hat\theta_k)^\top d_k + \tfrac12\Ov_k
+ \tfrac{\rho}{6}\norm{d_k}^3$; substitute the gradient identity and apply
the Cauchy--Schwarz inequality to the $\nabla J_k$ part.
(b) By the Cauchy--Schwarz inequality for the positive semi-definite form
$(x,y) \mapsto x^\top \sal_0\, y$,
$|\tau_k^\top \sal_0\, d_k| \le \norm{\tau_k}_{\sal_0}\norm{d_k}_{\sal_0}$.
The first factor is at most $\sqrt{2\Delta_k/\mu}$ by \eqref{eq:perbudget}
and the second at most $\sqrt{2/\mu}\,B_k(\bm{\lambda})$ by Step~2 of the
proof of Theorem~\ref{thm:ceiling}; multiply by $\mu$.
(c) Combine (a) and (b) with $\tfrac12\Ov_k \le \tfrac{c}{\mu}B_k^2$, which
is \eqref{eq:ceiling}.
\end{proof}

\begin{remark}[What the first-order term changes]
\label{rem:firstorder}
Three consequences follow. First, at an exact stationary point of $J_k$
the linear term of the expansion is exactly
$-\mu\,\tau_k^\top \sal_0\, d_k(\bm{\lambda})$. It is signed: at
$\bm{\lambda} = (1,\dots,1)$ it equals
$-\mu \sum_{j \ne k} \tau_k^\top \sal_0\, \tau_j$, so siblings positively
aligned with $\tau_k$ in the base metric lower the merged loss on task $k$
and conflicting siblings raise it. Second, the bound in (b) carries no
factor $1/\mu$; it shrinks with $\mu$ only through the descents
$\Delta_j$, whereas the curvature term carries $c/\mu$ explicitly. If
$\sal_0 \succ 0$ and the runs reach stationary points of the $J_j$, then
$\mu\,\sal_0\tau_j = -\nabla\loss_j(\hat\theta_j)$ gives
$\norm{\tau_j} = O(1/\mu)$ as $\mu \to \infty$, so the first-order term is
$O(1/\mu)$ while $\Ov_k$ is $O(1/\mu^2)$: for strong penalties the
first-order term generically dominates the excess loss and the ceiling of
Theorem~\ref{thm:ceiling} controls a vanishing share of it. Third,
composability under regularized training is therefore decided, to leading
order, by the base-metric alignment of the task vectors, which
single-task training does not control (compare
Remark~\ref{rem:ceilingloose}). What training enforces is control of the
curvature term, not composability as such. Figure~\ref{fig:penalty} (right) shows the crossover between the two terms numerically.
\end{remark}

\begin{remark}[The isotropic penalty satisfies the same theorem]
\label{rem:isotropic}
Theorem~\ref{thm:ceiling} does not by itself distinguish a
curvature-guided penalty from an isotropic one. With $\sal_0 = I$ the
objective \eqref{eq:regularized} is $L^2$-SP regularization,
Assumption~\ref{ass:domination} holds with $c = L$ by $L$-smoothness, and
\eqref{eq:ceiling} reads
$\Ov_k(\bm{\lambda}) \le \tfrac{2L}{\mu}B_k(\bm{\lambda})^2$. Because of
the scale invariance noted after Assumption~\ref{ass:domination}, the
ratios $c/\mu$ of two different statistics are comparable only at matched
own-task loss, which the theorem does not quantify. Whether curvature
guidance buys a better interference/specialization frontier than
isotropic shrinkage is thus an empirical question that the present theory
leaves open. Section~\ref{sec:exp_logistic} compares the two penalties on an exactly computable model.
\end{remark}

\begin{remark}[What the ceiling says, and what it costs]
\label{rem:ceilingreading}
Three features of \eqref{eq:ceiling} are worth isolating.

\emph{The $1/\mu$ scaling.} The ceiling is inversely proportional to
the penalty coefficient, so $\mu$ is a dial that trades merged-model
interference against per-task specialization, and
Theorem~\ref{thm:ceiling} makes the interference side of that trade
quantitative rather than empirical. Section~\ref{sec:exp_logistic} measures the constant $c$ and the looseness of the ceiling on an exactly computable model.

\emph{The role of $\Delta_j$.} The bound is driven by how far each task
descends from the base model, not by how far it \emph{moves}: a task
that reduces its loss a great deal is permitted a correspondingly larger
task vector, and contributes correspondingly more potential
interference. This is a natural accounting---a task vector that buys
no loss reduction should not be allowed to consume budget---and $\Delta_j$ is available at training time at the cost of evaluating $\loss_j$ at $\theta_0$ and at $\hat\theta_j$ (Algorithm~\ref{alg:mat}, line~12). Since $\loss_j$ is a population loss, in practice this is a held-out estimate and carries its sampling error.

\emph{The $\ell_1$ geometry.} In the simplified form
\eqref{eq:ceiling-simple} the dependence on the merging weights is
through $\norm{\bm{\lambda} - e_k}_1$, the distance from $\bm{\lambda}$
to the one-hot weighting that recovers task $k$ exactly. The ceiling
therefore degrades as the merge moves away from any single expert, at a
rate the penalty coefficient controls; at $\bm{\lambda} = e_k$ it is
zero, as it must be.

The cost of the theorem is Assumption~\ref{ass:domination} and the
constant $c$, which enters linearly. Because
Theorem~\ref{thm:ceiling} is a worst-case chain---a Loewner comparison
followed by a triangle inequality---$c$ and the bracket are both
pessimistic relative to realized interference; the theorem is a
guarantee, not an estimate, and $\widehat{\Ov}_k$ of
Remark~\ref{rem:computable} remains the quantity to report when an
estimate is what is wanted.
\end{remark}

\begin{remark}[No definiteness of $\Hess_k$ is required]
\label{rem:ceilingnodef}
The proof does not use $\Hess_k \succeq 0$. The comparison
$\Ov_k = d_k^\top \Hess_k d_k \le c\, d_k^\top \sal_0 d_k$ is a direct
consequence of Assumption~\ref{ass:domination} for any symmetric
$\Hess_k$, and the triangle inequality is applied in the $\sal_0$
seminorm, where positive semi-definiteness holds by construction. The
ceiling therefore applies at fine-tuned solutions that are only
approximate local minimizers, consistently with
Assumption~\ref{ass:stationary}, and does not inherit the
positive-semi-definiteness restriction of
Section~\ref{sec:characterization}.
\end{remark}

\begin{remark}[Factorized statistics and stochastic training]
\label{rem:ceilingfact}
Two extensions are immediate.

\emph{Factorization.} If training uses the rank-$r$ statistic
$\sal_0^{(r)}$ in place of $\sal_0$, as
Corollary~\ref{cor:factorized}(b) permits, then
Corollary~\ref{cor:factorized}(c) gives
$\norm{\tau_j}_{\sal_0}^2 \le \tfrac{2}{\mu}\Delta_j
+ \epsilon_r(\sal_0)\norm{\tau_j}^2$ for each $j$, and the proof of
Theorem~\ref{thm:ceiling} goes through with $\sqrt{2\Delta_j/\mu}$
replaced by $\bigl(\tfrac{2}{\mu}\Delta_j
+ \epsilon_r(\sal_0)\norm{\tau_j}^2\bigr)^{1/2}$ throughout, the
additional term being the measurable spectral tail already appearing in
Corollary~\ref{cor:factorized}.

\emph{Stochastic training.} The hypothesis
$J_j(\hat\theta_j) \le J_j(\theta_0)$ is a deterministic sublevel
condition on the returned iterate, not a statement about the dynamics
that produced it, so Theorem~\ref{thm:ceiling} applies pathwise to any
realization of stochastic training that satisfies it---and
Algorithm~\ref{alg:mat} verifies it directly, since line~12 returns
$\tfrac{2}{\mu}\Delta_j$ and line~8 monitors
$\Phi_t \le B_t$ along the way. This is why the theorem is stated for
the sublevel condition rather than in expectation: the certificate is
computed, so no distributional statement about the dynamics is needed. What remains is the estimation error of the two loss evaluations, since the sublevel condition concerns the population loss; and along a single stochastic run the monitored inequality $\Phi_t \le B_t$ is guaranteed by Theorem~\ref{thm:dynamics}(ii) only in expectation. If one prefers an
unconditional in-expectation form, Theorem~\ref{thm:dynamics}(ii)
supplies $\E\,\Phi(\theta^{(t)}) \le \tfrac{2}{\mu}
(\loss_j(\theta_0) - \E\loss_j(\theta^{(t)})) + \gamma^2 L_\mu\sigma^2/\mu$,
and the same algebra applies to the inflated budgets, at the cost of a
Jensen step when the square root is moved inside the expectation.
\end{remark}

\begin{remark}[Relation to proximal fine-tuning and continual learning]
With $\sal_0 = I$, \eqref{eq:regularized} is $L^2$-SP regularization
\citep{li2018explicit}; with $\sal_0$ the Fisher of \emph{previous
tasks}, it is elastic weight consolidation \citep{kirkpatrick2017overcoming},
of a piece with the importance-weighted penalties of the continual-learning
literature \citep{zenke2017continual, aljundi2018memory}. It is distinct
from decoupled weight decay \citep{loshchilov2019decoupled}, which pulls
toward the origin rather than toward $\theta_0$ in a curvature metric.
The novelty is not the penalty's form but its target and its cost: the
objective is forward-looking composability with \emph{unknown future
siblings} rather than retention of known past tasks, and $\sal_0$ can be obtained as a byproduct of the optimizer \citep{moayedikia2026bridging} rather than by a separate estimation pass (in Section~\ref{sec:transformer} it is nonetheless estimated separately).
Theorem~\ref{thm:dynamics}(ii)--(iii) together with
Theorem~\ref{thm:ceiling} is, to our knowledge, the first explicit link
from a training-time penalty coefficient to a merge-time interference
budget: Theorem~\ref{thm:dynamics}(ii) establishes control of the
surrogate not merely at convergence but uniformly along the entire
trajectory, and Theorem~\ref{thm:ceiling} converts that control into a
ceiling on the interference functional itself, conditional on Assumption~\ref{ass:domination} and leaving the first-order term of Proposition~\ref{prop:firstorder} outside its scope.
\end{remark}

\begin{corollary}[Factorized statistics suffice]\label{cor:factorized}
For a symmetric positive semi-definite statistic $\sal$, let
$\sal^{(r)}$ denote its best rank-$r$ approximation (spectral
truncation) and $\epsilon_r(\sal) = \sigma_{r+1}(\sal)$ the
truncation error. Then:
\begin{itemize}
\item[(a)] \emph{(Merge-time evaluation.)} The computed interference
functional $\widehat{\Ov}_k = \norm{d_k(\bm{\lambda})}_{\sal_k^{(r)}}^2$
satisfies
$\bigl|\widehat{\Ov}_k - \Ov_k(\bm{\lambda})\bigr| \le
\bigl(\delta + \epsilon_r(\sal_k)\bigr)\norm{d_k(\bm{\lambda})}^2$,
and consequently Theorem~\ref{thm:bound} holds with
$\widehat{\Ov}_k$ in place of $\Ov_k$ at the additive cost
$\tfrac12\bigl(\delta + \epsilon_r(\sal_k)\bigr)
\norm{d_k(\bm{\lambda})}^2$. Under
Assumptions~\ref{ass:rank}--\ref{ass:saliency},
$\epsilon_r(\sal_k) \le \delta + \min\{\eta, \tfrac{1}{r+1}\}
\operatorname{tr}(\Hess_k)$ when $\Hess_k \succeq 0$, the second
branch holding even without Assumption~\ref{ass:rank}.
\item[(b)] \emph{(Training.)} Theorem~\ref{thm:dynamics}(i)--(iii)
hold verbatim with $\sal_0^{(r)}$ in place of $\sal_0$ in the
objective \eqref{eq:regularized}; for parts (i) and (ii) even the
constants are unchanged, since $\norm{\sal_0^{(r)}}_{\mathrm{op}} =
\norm{\sal_0}_{\mathrm{op}}$ for every $r \ge 1$.
\item[(c)] \emph{(Budget transfer.)} If training uses
$\sal_0^{(r)}$, the interference budget of
Theorem~\ref{thm:dynamics}(iii) transfers to the full statistic:
any $\hat\theta_k$ with
$J_k^{(r)}(\hat\theta_k) \le J_k^{(r)}(\theta_0)$ satisfies
\[
\norm{\hat\theta_k - \theta_0}_{\sal_0}^2
\;\le\;
\frac{2}{\mu}\bigl(\loss_k(\theta_0) - \loss_k(\hat\theta_k)\bigr)
+ \epsilon_r(\sal_0)\,\norm{\hat\theta_k - \theta_0}^2 .
\]
\end{itemize}
\end{corollary}

\noindent See Appendix~\ref{app:factorized} for the proof. The
quantity $\epsilon_r(\sal_0)$ is directly measurable, and the
spectral measurements of \citep{moayedikia2026bridging}---stable
rank near $3$, over $82\%$ of energy at rank $32$---indicate it is
small at practically used ranks, which is what makes part (c) a
guarantee rather than a formality. The spectral bound in part (a) is weak when $\operatorname{tr}(\Hess_k)$ is large relative to $\norm{\Hess_k}_{\mathrm{op}}$, as is typical in high dimension; the measured $\epsilon_r$ should be reported in preference to it.

\section{Numerical Experiments}\label{sec:numerics}

Three studies are reported. The first two use multiclass logistic
regression, where losses, gradients and Hessians are exact, to measure how
tight the bound of Theorem~\ref{thm:bound} is outside the quadratic regime
and to test the statements of Section~\ref{sec:dynamics}, including the
constant $c$ of Assumption~\ref{ass:domination} and the first-order term
of Proposition~\ref{prop:firstorder}. The third asks whether the functional
retains predictive value at transformer scale, where the curvature
statistic is an optimizer byproduct rather than an exact Hessian. The
transformer study is illustrative rather than competitive; head-to-head
benchmarking of merge operators belongs to the companion methods paper.
All seeds are fixed, and the code and stored results for the three studies
accompany the paper.

\subsection{Tightness of the Bound Outside the Quadratic Regime}
\label{sec:exp_nonquad}

For quadratic losses $\tfrac12\Ov_k$ equals the excess loss identically
(Theorem~\ref{thm:characterization}), so a numerical check in that regime
tests only the implementation: over $2000$ random quadratics the two
quantities agree to $2.8\times10^{-14}$. The informative question is how
the bound of Theorem~\ref{thm:bound} behaves on a loss that is not
quadratic.

We take three-class logistic regression on four features plus a bias
($15$ parameters). A base model $\theta_0$ is trained on a pretraining
sample; two tasks are generated whose class means are a $0.6/0.4$ mixture
of the pretraining class means and task-specific ones. Each task is
trained to an exact minimizer $\theta_k$ by Newton's method, so that
$\epsilon_g \le 2.8\times10^{-14}$, and the merged model is
$\theta_k + s\,\tau_j$ for a scale $s \in [0.01, 2]$, where $s = 1$ is
task arithmetic with unit weights. For each $s$ we compute the exact
excess loss, the curvature term $\tfrac12\Ov_k$, and the full bound of
Theorem~\ref{thm:bound}, with $\rho$ estimated as the running maximum of the
Hessian difference quotient along the segment on a $60$-point grid. The
experiment is repeated over $20$ random problems and both tasks, giving
$40$ cases.

\begin{figure}[htbp]
\centering
\includegraphics[width=\textwidth]{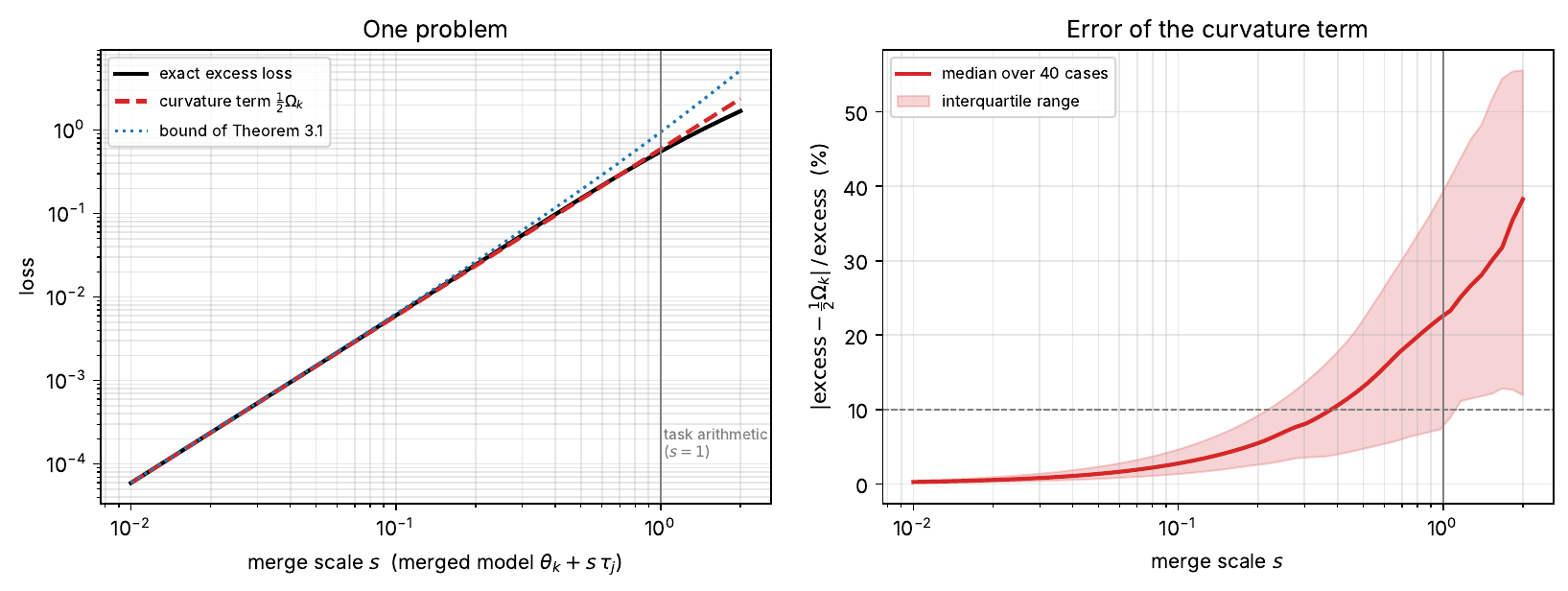}
\caption{Logistic regression, exact quantities. Left: for one problem,
the excess loss of the merged model $\theta_k + s\,\tau_j$, the curvature
term $\tfrac12\Ov_k$ and the bound of Theorem~\ref{thm:bound}, against
the merge scale $s$. Right: relative error of the curvature term as a
predictor of the excess loss, median and interquartile range over $40$
cases; the dashed line marks $10\%$.}
\label{fig:nonquad}
\end{figure}

That bound holds at every scale in every case. The
curvature term alone is an accurate predictor only locally
(Figure~\ref{fig:nonquad}): its median relative error is $2.9\%$ at
$s = 0.1$, $7.1\%$ at $s = 0.25$, $13.6\%$ at $s = 0.5$, $22\%$ at $s = 1$
and $38\%$ at $s = 2$, with $90$th percentiles of $6.3\%$, $16\%$, $34\%$,
$62\%$ and $69\%$. The median scale up to which it stays within $10\%$ of
the excess loss is $s = 0.36$. The error is not one-sided: $\tfrac12\Ov_k$
exceeds the excess loss in $23$--$33\%$ of cases, depending on the scale.
The full bound is looser than the excess loss by a median factor of $1.08$
at $s = 0.1$, $1.8$ at $s = 1$ and $2.5$ at $s = 2$. At unit weights,
then, the quadratic model is a roughly $20\%$ approximation on this
problem, and the exact criterion of Theorem~\ref{thm:characterization}
should be read as a local statement.

\subsection{Budget, Ceiling and Frontier on Logistic Regression}
\label{sec:exp_logistic}

The model is as in Section~\ref{sec:exp_nonquad}. The base model
$\theta_0$ is the minimizer of the loss on a pretraining sample of $600$
points, and the base statistic $\sal_0$ is the Hessian of that loss at
$\theta_0$, scaled to unit operator norm; it therefore contains no data
from the downstream tasks. It has rank $10$ of $15$ because of the
class-shift invariance of the softmax, a null space the task Hessians
share, so Assumption~\ref{ass:domination} is evaluated on the range of
$\sal_0$. Two tasks with $300$ samples each are fine-tuned from $\theta_0$
by gradient descent on \eqref{eq:regularized} with step size $1/L_\mu$,
where $L$ is taken as $1.5$ times the largest Hessian eigenvalue at
$\theta_0$, and merged as $\theta_0 + \tfrac12(\tau_A + \tau_B)$. Every
fine-tuning run is carried to stationarity of $J_k$ (gradient norm below
$10^{-13}$).

\begin{figure}[htbp]
\centering
\includegraphics[width=\textwidth]{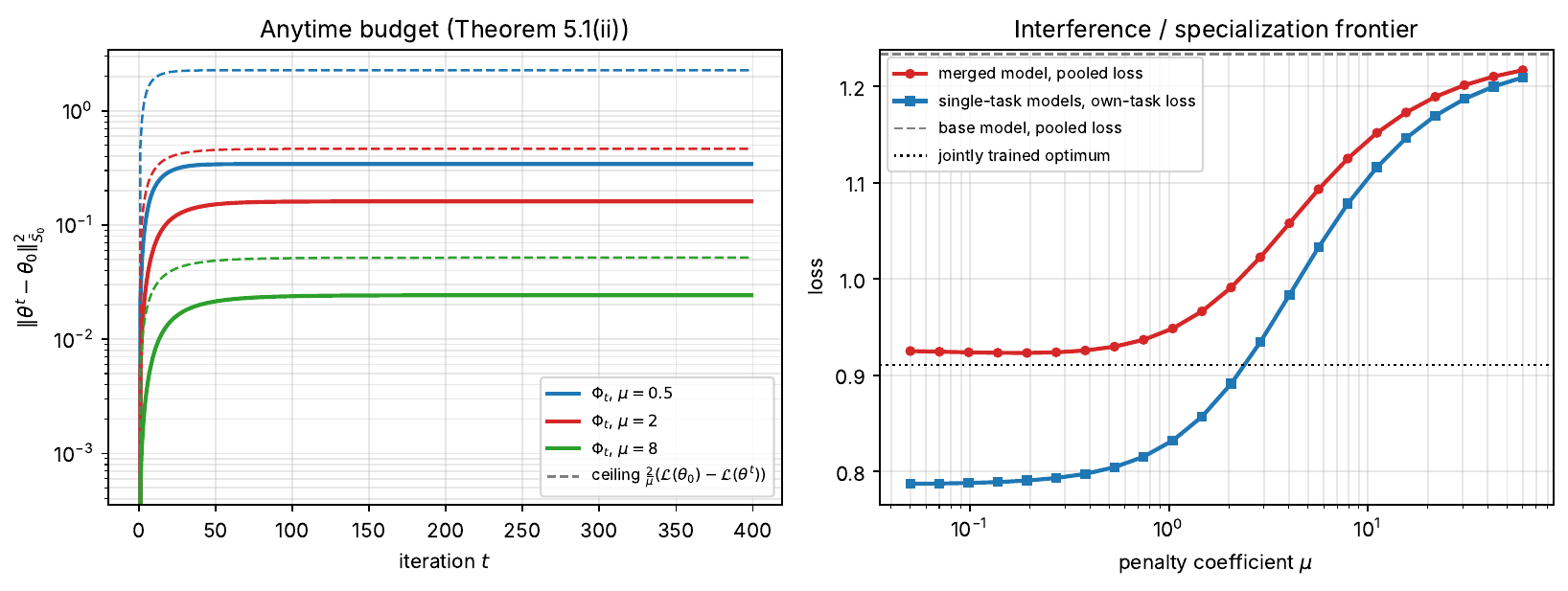}
\caption{Logistic regression, exact quantities. Left: the surrogate
$\Phi_t = \norm{\theta^t - \theta_0}_{\sal_0}^2$ (solid) stays below the
ceiling of Theorem~\ref{thm:dynamics}(ii) (dashed) at every iterate, for
three penalty coefficients. Right: pooled loss of the merged model and
own-task loss of the single-task models against $\mu$, with the base
model and the jointly trained optimum for reference.}
\label{fig:budget}
\end{figure}

\emph{Anytime budget.} For $\mu \in \{0.5, 2, 8\}$ and both tasks, the
maximum over iterates of $\Phi_t$ minus the ceiling of
Theorem~\ref{thm:dynamics}(ii) is zero, attained only at $t = 0$
(Figure~\ref{fig:budget}, left).

\emph{Frontier.} The pooled loss of the base model is $1.234$ and that of
a model trained jointly on both tasks is $0.911$. The merge of the
unregularized single-task models reaches $0.928$; the best merged model
under the penalty reaches $0.923$ at $\mu \approx 0.19$, at an own-task
loss of $0.791$ against $0.787$ without the penalty. For $\mu > 0.4$ the
merged loss rises again, as specialization is lost, and returns to the
base value for large $\mu$ (Figure~\ref{fig:budget}, right). Merging
therefore recovers most of the gap between the base model and the joint
optimum, but the penalty accounts for little of it: it improves the
unregularized merge by $0.5\%$. What the penalty does reduce is the excess
loss of the merged model relative to the single-task models, which falls
monotonically in $\mu$ (from $0.187$ at $\mu = 0.05$ to $0.011$ at
$\mu = 60$ for task $A$), and this is paid for in own-task loss.

\begin{figure}[htbp]
\centering
\includegraphics[width=\textwidth]{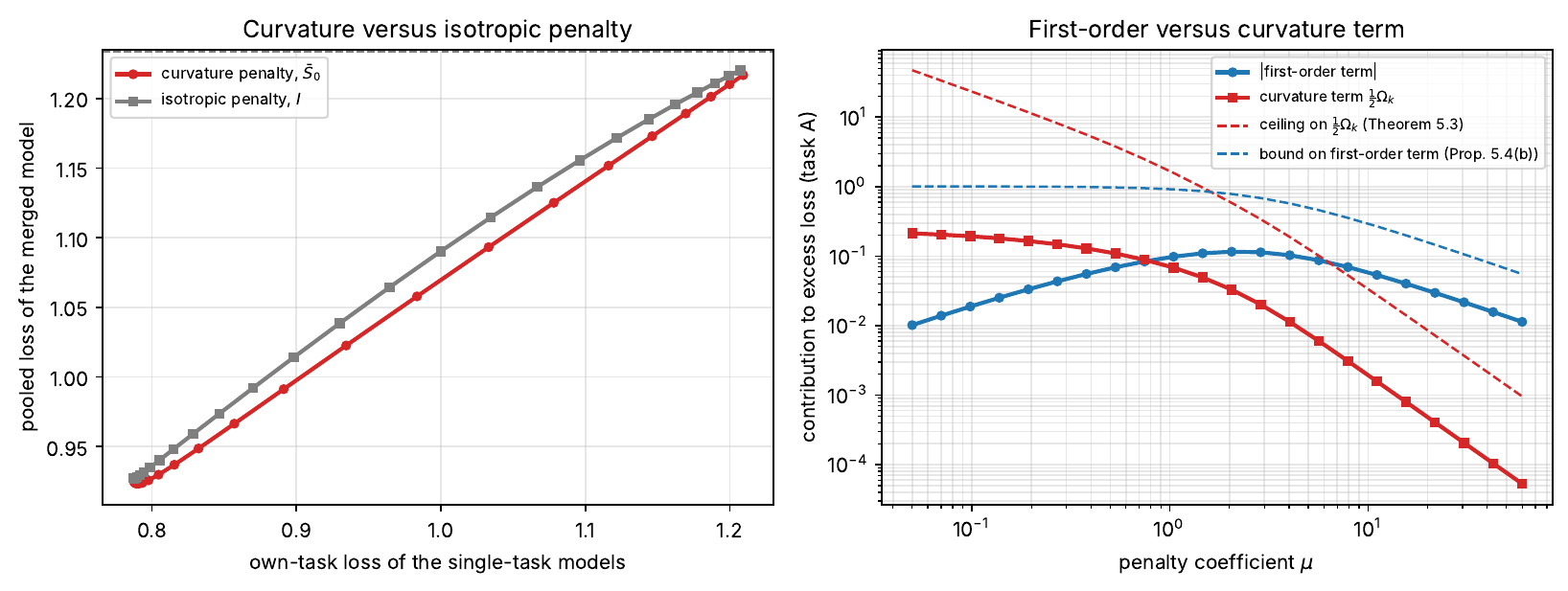}
\caption{Logistic regression, exact quantities. Left: pooled loss of the
merged model against the own-task loss of the single-task models, for the
curvature-guided penalty and the isotropic one; lower is better. Right:
the two contributions to the excess loss of the merged model on task $A$
against $\mu$, with the ceiling of Theorem~\ref{thm:ceiling} and the bound
of Proposition~\ref{prop:firstorder}(b) (dashed).}
\label{fig:penalty}
\end{figure}

\emph{Curvature versus isotropic penalty.} Repeating the sweep with
$\sal_0 = I$ and comparing at matched own-task loss, the curvature-guided
penalty gives a lower pooled loss of the merged model across the whole
frontier, by $0.014$ to $0.021$ (Figure~\ref{fig:penalty}, left). To check
that this is not specific to one problem we repeated the comparison on
$24$ problems, varying the class overlap, the relatedness of the tasks to
the pretraining classes, and the random draw. The curvature-guided penalty
was better in $20$ of the $24$; the mean advantage ranged from $0.012$
when the tasks share $60\%$ of their class structure with the pretraining
data to $0.001$ when they share $90\%$. The advantage is thus consistent
in sign but small, and it shrinks as the tasks become easier to merge.
This is the only evidence in the paper for curvature guidance over
isotropic shrinkage, and Remark~\ref{rem:isotropic} explains why the
theory does not supply any.

\emph{Ceiling.} The smallest constant satisfying
Assumption~\ref{ass:domination} lies between $2.4$ and $5.4$ over the
sweep. The ceiling \eqref{eq:ceiling} holds at every $\mu$ for both tasks
and is loose by a factor of $17$ for strong penalties, rising to $220$ at
$\mu = 0.05$ (Figure~\ref{fig:penalty}, right).

\emph{First-order term.} The gradient of the task loss at the regularized
solution grows from $0.013$ at $\mu = 0.05$ to $0.42$ at $\mu = 60$, as
Proposition~\ref{prop:firstorder}(a) requires. The first-order term of the
excess loss is smaller than the curvature term for weak penalties
($0.025$ against $0.180$ at $\mu = 0.14$), overtakes it between
$\mu = 0.4$ and $\mu = 1$, and is $5.7$ times larger at $\mu = 2.9$, $22$
times larger at $\mu = 7.9$ and more than $200$ times larger at
$\mu = 60$ (Figure~\ref{fig:penalty}, right). The bound of
Proposition~\ref{prop:firstorder}(b) holds throughout and is loose by a
factor of $5$ or more. In the range where the penalty is strong enough to
matter, then, the interference is governed by the first-order term and
not by $\Ov_k$, as Remark~\ref{rem:firstorder} anticipates.

\subsection{The Interference Functional at Transformer Scale}
\label{sec:transformer}

The third study asks whether the functional, computed from statistics the
optimizer already maintains, ranks merges by their observed degradation.
We fine-tune BERT-base on four GLUE tasks (RTE,
MRPC, CoLA, SST-2), forming five task pairs, with
AdamW for five epochs. The trajectory saliency $\bar S_k$ is the
bias-corrected Adam second-moment accumulator at the end of fine-tuning,
rescaled to unit operator norm; no curvature is computed separately for
the tasks. Each pair is merged with four operators (task arithmetic,
Fisher-weighted averaging, TIES, DARE) at two merge weights, $w = 0.5$ and
$w = 1$, giving $80$ merge evaluations. For each we compute the predicted
excess loss $\tfrac12\widehat{\Omega}_k
= \tfrac12\lVert d_k\rVert^2_{\bar S_k}$ from the actual displacement
$d_k = \theta_m - \theta_k$, the same quantity with $\bar S_k$ replaced by
its Adafactor row--column factorization, and, as a baseline that uses no
curvature at all, the squared Euclidean displacement $\lVert d_k\rVert^2$;
each is compared with the observed excess validation loss. The whole
experiment is repeated for three seeds, which differ in the initialization
of the classification heads and in the data order. The base statistic
$\sal_0$ used in the merge-aware runs is the diagonal Fisher information
of the pretrained model under its masked-language-modelling objective on
generic text; it is estimated by a separate pass and is therefore not an
optimizer byproduct in this study.

\emph{Predictiveness.} Table~\ref{tab:transformer} gives the Spearman rank
correlations with the observed excess loss. The functional is positively
correlated with observed degradation, $r_s = 0.30 \pm 0.08$ (mean and
standard deviation over seeds; $0.36$, $0.21$, $0.32$), and the factorized
statistic gives the same value although its relative truncation error
reaches $0.58$. The Euclidean baseline does better: $r_s = 0.39 \pm 0.04$,
higher than the functional in every seed, and higher within each of the
four operators. The two predictors are themselves strongly correlated
($0.87$ to $0.90$). Within a fixed merge weight the functional is not
significantly correlated with the outcome in any seed ($0.16 \pm 0.11$ at
$w = 0.5$ and $0.04 \pm 0.13$ at $w = 1$), so the pooled correlation
largely reflects the difference between the two weights: the mean observed
excess loss is $-0.01$ at $w = 0.5$ and $0.70$ at $w = 1$. The predicted
and observed magnitudes are not comparable, because $\bar S_k$ is
rescaled; the observed excess is about $2{,}500$ times the prediction
(median). In $73$ of the $240$ evaluations the merged model has a lower
validation loss than the single-task model, an outcome a positive
semi-definite quadratic model cannot represent and which reflects that the
fine-tuned models are not minimizers of the validation loss.

On this evidence, weighting the displacement by the optimizer's second
moment adds no predictive information beyond the length of the
displacement. The claim of Section~\ref{sec:characterization} that
composability is governed by the curvature metric rather than the
Euclidean one is a statement about the true Hessian; at transformer
scale, with this diagonal proxy, the experiment does not support it. A
diagonal second-moment statistic may simply be too poor a proxy for the
Hessian (Assumption~\ref{ass:saliency} with a large $\delta$), and we
cannot separate that explanation from a failure of the quadratic model
itself.

\begin{table}[htbp]
\centering
\caption{Transformer study: Spearman rank correlation between each
predictor and the observed excess loss, mean $\pm$ standard deviation over
three seeds ($80$ merge evaluations per seed; $40$ within each merge
weight and $20$ within each operator).}
\label{tab:transformer}
\small
\begin{tabular}{lcc}
\toprule
Subset & $\widehat{\Omega}_k$ (curvature-weighted) & $\lVert d_k\rVert^2$ (Euclidean) \\
\midrule
All evaluations & $0.30 \pm 0.08$ & $0.39 \pm 0.04$ \\
Merge weight $w = 0.5$ & $0.16 \pm 0.11$ & $0.28 \pm 0.07$ \\
Merge weight $w = 1$ & $0.04 \pm 0.13$ & $0.22 \pm 0.04$ \\
Task arithmetic & $0.27 \pm 0.18$ & $0.51 \pm 0.10$ \\
Fisher-weighted averaging & $0.53 \pm 0.05$ & $0.61 \pm 0.03$ \\
TIES & $0.47 \pm 0.09$ & $0.65 \pm 0.05$ \\
DARE & $0.38 \pm 0.22$ & $0.52 \pm 0.18$ \\
\bottomrule
\end{tabular}
\end{table}

\begin{figure}[htbp]
\centering
\includegraphics[width=\textwidth]{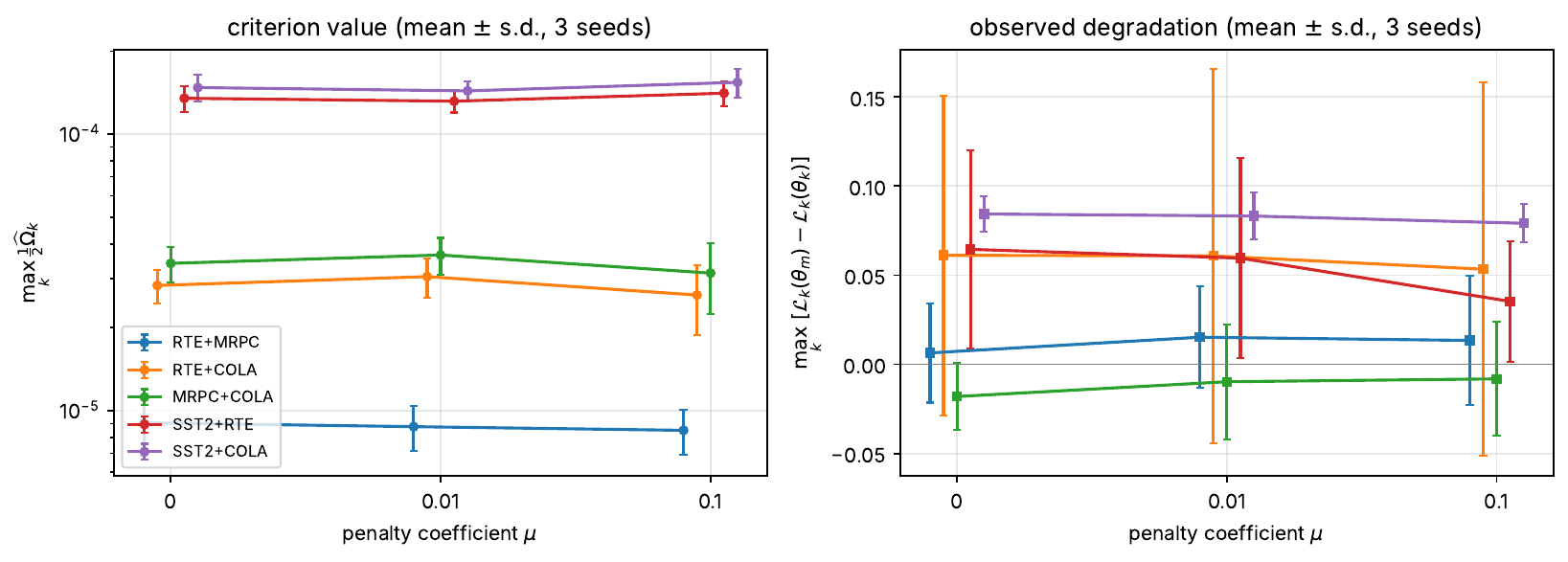}
\caption{Transformer study, merge-aware training (mean and standard
deviation over three seeds). Left: the criterion
$\max_k \tfrac12\widehat{\Omega}_k$ for each pair. Right: the observed
degradation. The penalty coefficients were too small to change the task
vectors, and no difference across $\mu$ exceeds the spread over seeds.}
\label{fig:regshift}
\end{figure}

\emph{Merge-aware training.} Every task was also fine-tuned under
\eqref{eq:regularized} with $\mu \in \{0.01, 0.1\}$ and the pairs merged
by task arithmetic at $w = 0.5$. With $\sal_0$ scaled to unit operator
norm these coefficients turned out too small to act: the squared norm of
every task vector changed by at most $0.5\%$ relative to $\mu = 0$. The
experiment is therefore not an informative test of the penalty. The
differences in observed degradation across $\mu$ are within the
seed-to-seed spread for every pair (Figure~\ref{fig:regshift}); only for
SST-2+RTE is the degradation lower at $\mu = 0.1$ than at $\mu = 0$ in all
three seeds ($0.064 \pm 0.055$ against $0.035 \pm 0.034$). The own-task
validation loss changed by between $-1.6\%$ and $+3.5\%$, with no
consistent sign. A meaningful test at this scale requires coefficients
large enough to shorten the task vectors, and we leave it to future work.

\begin{figure}[htbp]
\centering
\includegraphics[width=\textwidth]{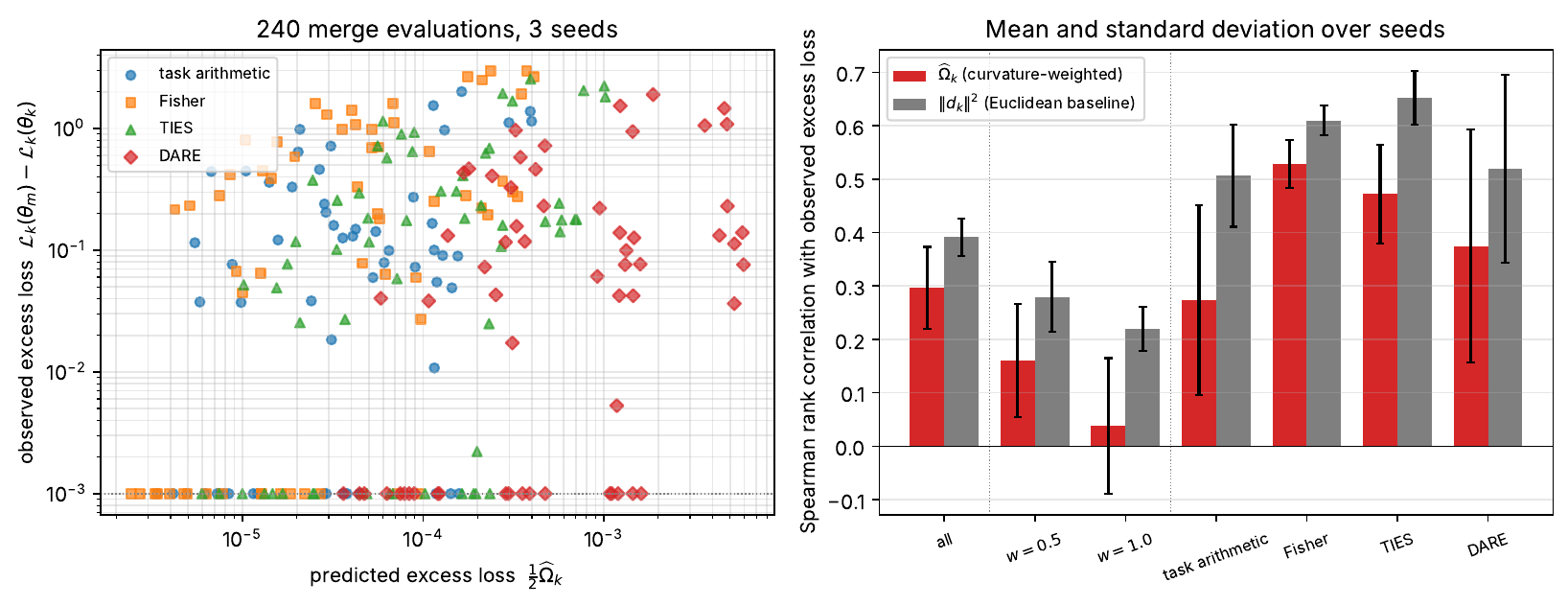}
\caption{Transformer study. Left: predicted against observed excess loss
for the $240$ merge evaluations of the three seeds; points on the dotted
line have non-positive observed excess. Right: Spearman rank correlation
with the observed excess loss for the curvature-weighted functional and
for the Euclidean baseline, over all evaluations, within each merge
weight and within each operator (mean and standard deviation over
seeds).}
\label{fig:omega_predicts}
\end{figure}

\section{Related Work}\label{sec:related}

The practice this paper analyzes sits within a large body of work on
combining models directly in weight space. Weight averaging as a route
to better solutions predates task arithmetic, from stochastic weight
averaging \citep{izmailov2018averaging} to model soups
\citep{wortsman2022soups}, and its feasibility rests on the linear mode
connectivity of networks trained from a shared basin
\citep{frankle2020linear, entezari2022role}. Task arithmetic
\citep{ilharco2023editing} established composition in weight space, and
the operators that followed can be read as different responses to the
same interference problem: Fisher-weighted averaging
\citep{matena2022fisher}, TIES \citep{yadav2023ties}, and DARE
\citep{yu2024dare} are the interference-suppression heuristics that
Remark~\ref{rem:operators} reads heuristically as acting on separate terms of
\eqref{eq:omega}, RegMean \citep{jin2023regmean} and AdaMerging
\citep{yang2024adamerging} fit closed-form or test-time coefficients,
and re-basin \citep{ainsworth2023gitrebasin} and activation-space
\citep{stoica2024zipit} methods first align models before merging, with
surveys \citep{yang2026model} documenting the resulting landscape.

These operators have in turn prompted a theoretical effort to explain
when composition succeeds, and it is against this effort that our
results are best positioned. \citet{ortizjimenez2023task} identify
weight disentanglement as the mechanism of task arithmetic and localize
it in the NTK regime; our Definition~\ref{def:composability} is the
weaker loss-level property, for which
Theorem~\ref{thm:characterization} gives a criterion that is also
necessary. Closer still, \citet{lu2026geometry} cast merging as a
subspace Fr\'echet mean under the Hessian metric and bound the
\emph{$\lambda$-weighted aggregate} loss gap
$\sum_t \lambda_t [\loss_t(\theta_m) - \loss_t(\theta_t)]$ by
Fr\'echet variance plus residual energy, with a characterization of
when curvature-aware merging beats flat-geometry merging, recovering
task arithmetic and Fisher averaging as Fr\'echet means under different
metrics. Our Theorem~\ref{thm:bound} differs from this in three
respects: it is \emph{per-task} (worst-case over $k$, which is what
Definition~\ref{def:composability} requires---an aggregate bound
permits sacrificing one task for the average), it decomposes the
quadratic form into operator-interpretable terms including the signed
alignment that sign-election methods target, and it is stated for
optimizer-byproduct statistics (Remark~\ref{rem:computable}); their
question of which \emph{operator} is optimal for a given collection is
complementary to ours, which asks when \emph{any} composition can
succeed and how training shapes the answer. A related
stability-theoretic framework \citep{li2026unified} decouples the
optimization and generalization error of merged models and derives
scaling laws predicting when experts are ``merge-friendly,'' but that
analysis takes the fine-tuning procedure as given and characterizes its
hyperparameters, whereas Theorem~\ref{thm:dynamics} modifies the
procedure itself and provides per-collection rather than on-average
guarantees, so the two are complementary and neither implies the other;
in the same vein, \citet{zhou2024atm} show that iterating tuning and
merging minimizes an upper bound on the joint fine-tuning loss, which is
the multi-round counterpart to our single-pass setting where no
merge-round iteration or cross-task data access is available.

Where the works above treat merging as an operation on already-trained
models, a smaller line intervenes during training, and it is here that
our Section~\ref{sec:dynamics} has its closest neighbor. The
KFAC-regularized fine-tuning of \citet{porrello2026dataless} promotes
function-space weight disentanglement during linearized fine-tuning
through a Jacobian--Gram penalty, made dataless by a curvature
approximation. Our penalty differs from it on three counts that recur
throughout the paper: it is designed for statistics the optimizer already
maintains, needing no KFAC estimation pass and remaining compatible
with memory-efficient training (in our transformer study the base statistic was nonetheless estimated by a separate diagonal-Fisher pass, cheaper than KFAC but not free); its guarantees are loss-level, with the
explicit convergence and budget control of
Theorem~\ref{thm:dynamics}(i) and (iii) where theirs are
representation-level; and its analysis extends to factorized statistics
(Corollary~\ref{cor:factorized}). The penalty's form is itself
inherited, tracing back to $L^2$-SP \citep{li2018explicit} and elastic
weight consolidation \citep{kirkpatrick2017overcoming}, with the
reframing from retention of past tasks to prospective composability
discussed after Theorem~\ref{thm:dynamics}; the empirical finding that
even generic regularization modestly improves mergeability
\citep{demystifying2026} is what motivates sharpening it into the
curvature-guided form analyzed here.

The machinery behind that analysis, finally, is standard optimization
theory. The convergence guarantee of Theorem~\ref{thm:dynamics}(i)
reproduces the classical $O(1/\sqrt{T})$ stationarity rate for smooth
non-convex stochastic optimization \citep{ghadimi2013stochastic,
bottou2018optimization}, specialized to the regularized objective, and
its adaptive-method reading follows the continuous-time and convergence
analyses of Adam and its relatives \citep{kingma2015adam,
reddi2018convergence, defossez2022simple, barakat2021convergence}, while
the Polyak--{\L}ojasiewicz condition \citep{karimi2016linear} is what
would upgrade the anytime budget of part (ii) to a rate. The
factorized-statistics setting connects to sublinear-memory optimizers
such as Adafactor \citep{shazeer2018adafactor} and extends
\citep{moayedikia2026bridging}, whose empirical spectral analysis
supplies the evidence for Assumption~\ref{ass:rank}.

\section{Conclusion}\label{sec:conclusion}

This paper has treated the composability of fine-tuned
models---whether their task vectors can be merged without sacrificing
per-task performance---as a quantity that can be measured before
merging from statistics an optimizer already maintains, and controlled
during training by acting on those same statistics. The excess
loss incurred by merging is controlled by an interference functional of
the cross-task curvature overlap and the signed alignment of the task
vectors (Theorem~\ref{thm:bound}); in the linearized regime this
functional does not merely bound but equals the excess loss, giving a
criterion that is necessary as well as sufficient
(Theorem~\ref{thm:characterization}) and that measures angle in the
curvature metric rather than the Euclidean one. Because the functional
is assembled from statistics an adaptive optimizer already maintains, it
can be evaluated before merging without data or a separate estimation
pass, and the same statistics let training be shaped for composition:
fine-tuning under a curvature-guided penalty keeps displacement in the
base curvature metric below a ceiling set by the penalty coefficient at
every iterate, not only at convergence (Theorem~\ref{thm:dynamics}),
while retaining the standard non-convex convergence rate and surviving
low-rank factorization of the curvature statistic
(Corollary~\ref{cor:factorized}). A single Loewner comparison between
the base statistic and the task curvatures
(Assumption~\ref{ass:domination}) converts that trajectory-level control
into a ceiling on the interference functional itself
(Theorem~\ref{thm:ceiling}), proportional to $1/\mu$ and to the
task-loss descent each single-task run achieves, so that the curvature part of the certificate
required before merging is assembled from outputs the training
runs return, together with the comparison constant $c$. On an exactly computable model the budget and the ceiling hold, the measured comparison constant is between $2.4$ and $5.4$, and the curvature-guided penalty gives a slightly better frontier than an isotropic one. The transformer study is less favourable: the functional is rank-correlated with merging degradation ($0.30 \pm 0.08$ over three seeds) but does no better than the Euclidean displacement norm when the
curvature is a genuine optimizer byproduct rather than an exact Hessian.

Several limitations delimit these claims. The necessity half of the
characterization holds in the quadratic regime; outside it only the
upper bound survives, and Remark~\ref{rem:cubic} shows the converse can
fail once third-order terms are admitted. The anytime budget is stated
for the case in which the penalty gradient is exact and the
task-loss oracle carries the stochasticity, and the interference-decrease
identity is descriptive rather than a monotonicity guarantee. The
merge-time ceiling of Theorem~\ref{thm:ceiling} rests on
Assumption~\ref{ass:domination}, whose constant $c$ enters linearly and
is not assumed small; the ceiling is moreover worst-case in the sibling
alignment, since the triangle inequality of its proof discards the very
signed-alignment structure that Theorem~\ref{thm:bound} identifies as
the source of interference (Remark~\ref{rem:ceilingloose}). It is a
certificate that holds for any collection produced this way, not an
estimate of the interference a typical collection will exhibit. The
transformer study is deliberately illustrative: it validates the
functional as an ordering of merge quality, not as a numerical estimate
of excess loss, and it makes no competitive claim against merging
operators. Further limits should be stated. First, the ceiling controls only the curvature term: a regularized solution is not stationary for the task loss, and the resulting first-order term (Proposition~\ref{prop:firstorder}) is not reduced by the factor $1/\mu$ and dominates for strong penalties. Second, the theory does not separate the curvature-guided penalty from an isotropic one, which satisfies Theorem~\ref{thm:ceiling} with $c = L$ (Remark~\ref{rem:isotropic}). Third, the quadratic model is local: on logistic regression the curvature term is within $10\%$ of the excess loss only up to about a third of the unit merge weight. Fourth, at transformer scale the curvature-weighted functional did not outperform the Euclidean displacement norm, so the experiments do not support the metric claim there; and the merge-aware training runs used penalty coefficients too small to change the task vectors, so the effect of the penalty at that scale remains untested. The constant $c$ was measured only on the small model.

That competitive question---whether training for composability yields
merged models that outperform post-hoc operators across benchmarks, and
at what cost in per-task accuracy---is the subject of a companion
methods paper, for which the interference functional and the
curvature-guided penalty developed here provide the analytical
foundation. Beyond it, the results suggest treating the optimizer's
curvature trajectory as a first-class object: if the same statistics
that guide efficient training also determine merging quality, then the
design of optimizers, of regularizers, and of merging operators is
better approached jointly than in isolation.

\appendix
\section{Proofs}\label{app:proofs}

We first record an algebraic identity used throughout.

\begin{lemma}[Interference identity]\label{lem:identity}
For every $k \in [K]$ and all $\bm{\lambda}$,
$\Ov_k(\bm{\lambda}) = \norm{d_k(\bm{\lambda})}_{\Hess_k}^2$, where
$d_k(\bm{\lambda})$ is the displacement \eqref{eq:displacement} and
$\Ov_k$ is defined in \eqref{eq:omega}. In particular, if
$\Hess_k \succeq 0$ then $\Ov_k(\bm{\lambda}) \ge 0$.
\end{lemma}

\begin{proof}
Write $u = (\lambda_k - 1)\tau_k$ and
$v = \sum_{j \ne k} \lambda_j \tau_j$, so that
$d_k(\bm{\lambda}) = u + v$ by \eqref{eq:displacement}. Bilinearity of
$(x,y) \mapsto x^\top \Hess_k y$ and symmetry of $\Hess_k$ give
\[
\norm{u + v}_{\Hess_k}^2
= \norm{v}_{\Hess_k}^2 + 2\, u^\top \Hess_k v + \norm{u}_{\Hess_k}^2 ,
\]
and the three terms are, in order, the cross-task curvature overlap,
the signed alignment term, and the shrinkage term of
\eqref{eq:omega}. Nonnegativity for $\Hess_k \succeq 0$ is immediate
since $\Ov_k = \norm{d_k}_{\Hess_k}^2$.
\end{proof}

\subsection{Proof of Theorem~\ref{thm:bound}}\label{app:bound}

\begin{proof}
Fix $k \in [K]$ and abbreviate $\bar d = d_k(\bm{\lambda})$, so that
$\merged = \theta_k + \bar d$. By Assumption~\ref{ass:smooth} there is a
convex neighborhood $\mathcal{N}$ containing $\theta_k$ and $\merged$
on which $\loss_k$ is twice continuously differentiable with
$\rho$-Lipschitz Hessian; convexity of $\mathcal{N}$ ensures the
segment $\{\theta_k + s \bar d : s \in [0,1]\}$ lies in $\mathcal{N}$, so
all evaluations below are licensed.

Since $\loss_k$ is twice continuously differentiable on
$\mathcal{N}$, Taylor's theorem with integral remainder gives
\begin{equation}\label{eq:taylor}
\loss_k(\theta_k + \bar d)
= \loss_k(\theta_k)
+ \nabla \loss_k(\theta_k)^\top \bar d
+ \tfrac{1}{2}\, \bar d^\top \Hess_k\, \bar d
+ \mathcal{R},
\end{equation}
where
\[
\mathcal{R}
= \int_0^1 (1 - s)\; \bar d^\top
\bigl[\nabla^2 \loss_k(\theta_k + s \bar d) - \Hess_k\bigr]\, \bar d \; ds .
\]
We bound the three non-constant terms of \eqref{eq:taylor} in turn.

\emph{Linear term.} By the Cauchy--Schwarz inequality and
Assumption~\ref{ass:stationary},
\[
\nabla \loss_k(\theta_k)^\top \bar d
\;\le\; \norm{\nabla \loss_k(\theta_k)}\,\norm{\bar d}
\;\le\; \epsilon_g \norm{\bar d}.
\]

\emph{Quadratic term.} By Lemma~\ref{lem:identity},
$\bar d^\top \Hess_k \bar d = \norm{\bar d}_{\Hess_k}^2 = \Ov_k(\bm{\lambda})$, so the
quadratic term equals $\tfrac12 \Ov_k(\bm{\lambda})$ exactly.

\emph{Remainder.} For each $s \in [0,1]$, the $\rho$-Lipschitz
property of the Hessian on $\mathcal{N}$ gives
$\norm{\nabla^2 \loss_k(\theta_k + s \bar d) - \Hess_k}_{\mathrm{op}}
\le \rho\, s \norm{\bar d}$, hence
\[
\bigl| \bar d^\top [\nabla^2 \loss_k(\theta_k + s \bar d) - \Hess_k]\, \bar d \bigr|
\;\le\; \rho\, s\, \norm{\bar d}^3 ,
\]
and therefore
\[
|\mathcal{R}|
\;\le\; \rho \norm{\bar d}^3 \int_0^1 s(1 - s)\, ds
\;=\; \frac{\rho}{6} \norm{\bar d}^3 .
\]

Combining the three bounds in \eqref{eq:taylor} yields
\[
\loss_k(\merged) - \loss_k(\theta_k)
\;\le\; \tfrac{1}{2}\,\Ov_k(\bm{\lambda})
+ \epsilon_g \norm{\bar d}
+ \tfrac{\rho}{6} \norm{\bar d}^3 ,
\]
which is the claimed inequality. The composability consequence is
immediate: if the right-hand side is at most $\varepsilon$ for every
$k$, then $\loss_k(\merged) \le \loss_k(\theta_k) + \varepsilon$ for
every $k$, which is Definition~\ref{def:composability}.
\end{proof}

\begin{remark}[The bound is exact for quadratics]\label{rem:tightproof}
If $\loss_k$ is quadratic with Hessian $\Hess_k$ and
$\nabla \loss_k(\theta_k) = 0$, then $\rho = 0$ and $\epsilon_g = 0$
may be taken in Assumptions~\ref{ass:smooth}--\ref{ass:stationary},
the remainder $\mathcal{R}$ and the linear term vanish identically,
and \eqref{eq:taylor} reads
$\loss_k(\merged) - \loss_k(\theta_k) = \tfrac12 \Ov_k(\bm{\lambda})$.
Theorem~\ref{thm:bound} therefore holds with equality, so no bound of
this form can be uniformly sharper without additional assumptions on
$\loss_k$ beyond Assumptions~\ref{ass:smooth}--\ref{ass:stationary}.
\end{remark}

\subsection{Proof of Theorem~\ref{thm:characterization}}
\label{app:characterization}

\begin{proof}
By hypothesis each task loss is exactly quadratic,
\begin{equation}\label{eq:quadloss}
\loss_k(\theta)
= \loss_k(\theta_k) + \tfrac{1}{2}\norm{\theta - \theta_k}_{\Hess_k}^2,
\qquad \Hess_k \succeq 0,
\end{equation}
where positive semi-definiteness of $\Hess_k$ holds because
\eqref{eq:quadloss} attains its minimum at $\theta_k$. Evaluating
\eqref{eq:quadloss} at $\theta = \merged = \theta_k +
d_k(\bm{\lambda})$ and applying Lemma~\ref{lem:identity},
\begin{equation}\label{eq:exact}
\loss_k(\merged) - \loss_k(\theta_k)
= \tfrac{1}{2}\norm{d_k(\bm{\lambda})}_{\Hess_k}^2
= \tfrac{1}{2}\,\Ov_k(\bm{\lambda})
\qquad \text{for every } k \in [K].
\end{equation}
This identity is exact; no remainder or gradient term appears.

($\Leftarrow$) Suppose $\max_k \Ov_k(\bm{\lambda}) \le 2\varepsilon$.
Then by \eqref{eq:exact}, for every $k$,
$\loss_k(\merged) - \loss_k(\theta_k) = \tfrac12 \Ov_k(\bm{\lambda})
\le \varepsilon$, so the collection is $\varepsilon$-composable at
$\bm{\lambda}$.

($\Rightarrow$) Suppose the collection is $\varepsilon$-composable at
$\bm{\lambda}$. Then for every $k$,
$\loss_k(\merged) - \loss_k(\theta_k) \le \varepsilon$, and by
\eqref{eq:exact} this reads $\tfrac12 \Ov_k(\bm{\lambda}) \le
\varepsilon$ for every $k$, i.e.\
$\max_k \Ov_k(\bm{\lambda}) \le 2\varepsilon$.
\end{proof}

\begin{remark}[Necessity fails outside the quadratic regime]
\label{rem:cubic}
The two directions use the same identity \eqref{eq:exact}; the
theorem's content is that in the quadratic regime the interference
functional does not merely control the excess loss but \emph{is} the
excess loss (up to the factor $\tfrac12$). Outside the quadratic
regime the forward direction survives as Theorem~\ref{thm:bound}
while the converse fails: a large $\Ov_k$ can be canceled by
higher-order terms. A one-dimensional construction suffices. Let
$K = 2$, $d = 1$, and on a neighborhood of $\theta_1$ let
\[
\loss_1(\theta)
= \tfrac{h}{2}(\theta - \theta_1)^2 + \beta\,(\theta - \theta_1)^3 ,
\qquad h > 0 ,
\]
with the merging weights $\bm{\lambda} = (1,1)$ producing displacement
$d_1 = s > 0$ (e.g.\ $\tau_2 = s$). Then
$\loss_1(\theta_1 + s) - \loss_1(\theta_1) = \tfrac{h}{2}s^2
+ \beta\,s^3$ while $\Ov_1 = hs^2$. Given any target
$\varepsilon > 0$ and any $hs^2 > 2\varepsilon$, choosing
$\beta = (\varepsilon - \tfrac{h}{2}s^2)/s^3 < 0$ makes the
excess loss exactly $\varepsilon$ while $\Ov_1 = hs^2$ is
arbitrarily large: the pair is $\varepsilon$-composable, yet the
criterion of Theorem~\ref{thm:characterization} fails. The
construction sits inside
Assumptions~\ref{ass:smooth}--\ref{ass:stationary}: $\theta_1$
remains a strict local minimizer, since
$\loss_1'(\theta) = (\theta - \theta_1)\bigl[h + 3\beta(\theta -
\theta_1)\bigr] > 0$ for $\theta - \theta_1 \in (0, h/(3|\beta|))$, and
the Hessian is Lipschitz with $\rho = 6|\beta|$. This is why necessity is
claimed only under the quadratic restriction.
\end{remark}

\subsection{Computation for Example~\ref{ex:tight}}\label{app:example}

\begin{proof}[Computation]
Let $d = 2$, $K = 2$, $\Hess_1 = \Hess_2 = \mathrm{diag}(h, h')$ with
$h, h' > 0$, $\tau_1 = (a, 0)$, $\tau_2 = (b\cos\phi, b\sin\phi)$,
and $\lambda_1 = \lambda_2 = 1$. From \eqref{eq:displacement},
$d_1 = \tau_2$ and $d_2 = \tau_1$. By Lemma~\ref{lem:identity},
\[
\Ov_1 = \norm{\tau_2}_{\Hess_1}^2
= b^2\bigl(h\cos^2\phi + h'\sin^2\phi\bigr),
\qquad
\Ov_2 = \norm{\tau_1}_{\Hess_2}^2 = a^2 h .
\]
By Theorem~\ref{thm:characterization}, the pair is
$\varepsilon$-composable at $(1,1)$ if and only if
\begin{equation}\label{eq:phase}
b^2\bigl(h\cos^2\phi + h'\sin^2\phi\bigr) \le 2\varepsilon
\quad\text{and}\quad
a^2 h \le 2\varepsilon .
\end{equation}
Condition \eqref{eq:phase} defines the phase boundary in the
$(b, \phi)$ plane referred to in the example; it depends on $\phi$
only through the curvature-weighted quantity
$h\cos^2\phi + h'\sin^2\phi$, not the Euclidean angle alone.

\emph{Euclidean orthogonality is not sufficient.} Take
$\phi = \pi/2$, so $\tau_1^\top \tau_2 = ab\cos\phi = 0$. Then
$\Ov_1 = b^2 h'$, and for any fixed $\varepsilon, b > 0$, choosing
$h' > 2\varepsilon / b^2$ violates \eqref{eq:phase}: the pair is
Euclidean-orthogonal yet not $\varepsilon$-composable. ($\Ov_2 = a^2 h$ is controlled by taking $a$ small.)

\emph{Euclidean orthogonality is not necessary.} Take $\phi = 0$, so
$\tau_1^\top \tau_2 = ab \ne 0$ for $a, b > 0$: the task vectors are
parallel, the maximal Euclidean overlap. Then
$\Ov_1 = b^2 h$ and $\Ov_2 = a^2 h$, and choosing
$h \le 2\varepsilon / \max(a^2, b^2)$ (and any $h' > 0$) satisfies
\eqref{eq:phase}: the pair composes freely despite full Euclidean
overlap.

Finally, by Remark~\ref{rem:tightproof} the losses here are exactly
quadratic, so all statements hold with equality and the example
simultaneously exhibits tightness of Theorem~\ref{thm:bound}.
\end{proof}

\subsection{Proof of Theorem~\ref{thm:dynamics}(i)}\label{app:sgd}

We first record the smoothness of the regularized objective.

\begin{lemma}[Smoothness of $J_k$]\label{lem:smoothJ}
Under Assumptions~\ref{ass:smooth} and \ref{ass:saliency}, the
objective $J_k$ of \eqref{eq:regularized} is $L_\mu$-smooth on
$\mathcal{N}$ with $L_\mu = L + \mu\norm{\sal_0}_{\mathrm{op}}$; that
is, $\norm{\nabla J_k(\theta) - \nabla J_k(\theta')} \le L_\mu
\norm{\theta - \theta'}$ for all $\theta, \theta' \in \mathcal{N}$.
\end{lemma}

\begin{proof}
$\nabla J_k(\theta) = \nabla \loss_k(\theta) + \mu \sal_0 (\theta -
\theta_0)$. For $\theta, \theta' \in \mathcal{N}$,
\[
\norm{\nabla J_k(\theta) - \nabla J_k(\theta')}
\le \norm{\nabla \loss_k(\theta) - \nabla \loss_k(\theta')}
+ \mu \norm{\sal_0 (\theta - \theta')}
\le \bigl(L + \mu \norm{\sal_0}_{\mathrm{op}}\bigr)
\norm{\theta - \theta'},
\]
using $L$-smoothness of $\loss_k$ (Assumption~\ref{ass:smooth}) and
the definition of the operator norm.
\end{proof}

\begin{proof}[Proof of Theorem~\ref{thm:dynamics}(i)]
Throughout, $\mathcal{F}_t$ denotes the $\sigma$-algebra generated by
$g_0, \dots, g_t$, expectations without subscript are total, and we
abbreviate $J = J_k$, $\theta^t = \theta^{(t)}$,
$\nabla J^t = \nabla J(\theta^{(t)})$. The update is
$\theta^{t+1} = \theta^t - \gamma_T\, g_t$ with
$\gamma_T = \gamma/\sqrt{T}$ fixed for the whole run.

\emph{Step 1: descent inequality.} By Lemma~\ref{lem:smoothJ} and the
standard descent lemma for $L_\mu$-smooth functions,
\[
J(\theta^{t+1})
\;\le\;
J(\theta^t)
- \gamma_T \,\langle \nabla J^t,\, g_t \rangle
+ \frac{L_\mu \gamma_T^2}{2} \norm{g_t}^2 .
\]
Taking conditional expectation given $\mathcal{F}_{t-1}$ and using
unbiasedness, $\E[\langle \nabla J^t, g_t\rangle \mid
\mathcal{F}_{t-1}] = \norm{\nabla J^t}^2$, while the bounded-variance
condition and orthogonality of the noise give
\[
\E\bigl[\norm{g_t}^2 \mid \mathcal{F}_{t-1}\bigr]
= \norm{\nabla J^t}^2
+ \E\bigl[\norm{g_t - \nabla J^t}^2 \mid \mathcal{F}_{t-1}\bigr]
\le \norm{\nabla J^t}^2 + \sigma^2 .
\]
Combining and taking total expectations,
\begin{equation}\label{eq:onestep}
\E\, J(\theta^{t+1})
\;\le\;
\E\, J(\theta^t)
- \gamma_T \Bigl(1 - \frac{L_\mu \gamma_T}{2}\Bigr)
\E \norm{\nabla J^t}^2
+ \frac{L_\mu \gamma_T^2 \sigma^2}{2} .
\end{equation}

\emph{Step 2: step size condition.} The hypothesis
$T \ge \gamma^2 L_\mu^2$ gives $\gamma_T = \gamma/\sqrt{T} \le
1/L_\mu$, hence $1 - L_\mu \gamma_T / 2 \ge 1/2$, and
\eqref{eq:onestep} implies
\[
\frac{\gamma_T}{2}\, \E \norm{\nabla J^t}^2
\;\le\;
\E\, J(\theta^t) - \E\, J(\theta^{t+1})
+ \frac{L_\mu \gamma_T^2 \sigma^2}{2} .
\]

\emph{Step 3: telescoping.} Summing over $t = 0, \dots, T-1$, the
middle terms telescope:
\[
\frac{\gamma_T}{2} \sum_{t=0}^{T-1} \E \norm{\nabla J^t}^2
\;\le\;
J(\theta^0) - \E\, J(\theta^T)
+ \frac{L_\mu \gamma_T^2 \sigma^2}{2}\, T .
\]

\emph{Step 4: bounding the initial gap.} Because the iterate sequence
is initialized at the base model, $\theta^0 = \theta_0$, the
regularizer vanishes there and $J(\theta^0) = \loss_k(\theta_0)$.
Because $\sal_0 \succeq 0$ (Assumption~\ref{ass:saliency}), the
regularizer is nonnegative everywhere, so
$J(\theta) \ge \loss_k(\theta) \ge \loss_k^\star$ for all $\theta$,
whence $\E\, J(\theta^T) \ge \loss_k^\star$. Therefore
\[
J(\theta^0) - \E\, J(\theta^T)
\;\le\; \loss_k(\theta_0) - \loss_k^\star .
\]

\emph{Step 5: conclusion.} Dividing by $\gamma_T T / 2$ and using
$\min_t \le$ average,
\[
\min_{0 \le t < T} \E \norm{\nabla J^t}^2
\;\le\;
\frac{2\bigl(\loss_k(\theta_0) - \loss_k^\star\bigr)}{\gamma_T\, T}
+ L_\mu \gamma_T \sigma^2
\;=\;
\frac{2\bigl(\loss_k(\theta_0) - \loss_k^\star\bigr)}{\gamma \sqrt{T}}
+ \frac{L_\mu \gamma \sigma^2}{\sqrt{T}} ,
\]
where the equality substitutes $\gamma_T = \gamma/\sqrt{T}$. This is
the claimed bound, and since $L_\mu = L + \mu\norm{\sal_0}_{\mathrm{op}}$
with $L$, $\gamma$, $\sigma$ fixed, the right-hand side is
$O\bigl((1 + \mu\norm{\sal_0}_{\mathrm{op}})/\sqrt{T}\bigr)$.
\end{proof}

\begin{remark}[On the hypotheses, and the price of the regularizer]
\label{rem:sgdhyp}
Two hypotheses deserve comment. First, the requirement that
Assumption~\ref{ass:smooth} hold on a set containing the iterates is
the standard stability caveat of non-convex SGD analysis; it is
satisfied trivially when $\loss_k$ is smooth on all of $\R^d$, and
can otherwise be enforced, e.g., by projection onto a compact set, at
the cost of a standard modification of the analysis
\citep{barakat2021convergence}. Second, the noise model places
the stochasticity on the oracle for $\nabla J_k$; in practice only
$\nabla \loss_k$ is sampled while the regularizer's gradient
$\mu\sal_0(\theta - \theta_0)$ is computed exactly, which is a
special case. The theorem quantifies the price of merge-awareness in
optimization terms: the regularizer inflates the smoothness constant
from $L$ to $L + \mu\norm{\sal_0}_{\mathrm{op}}$, degrading the
constant---not the rate---and the degradation is controlled by the
same knob $\mu$ that buys the interference budget of part (iii).
\end{remark}

\subsection{Proof of Theorem~\ref{thm:dynamics}(ii) and (iii)}
\label{app:budget}

Both parts rest on a single rearrangement, which we isolate.

\begin{lemma}[Budget from sublevel membership]\label{lem:budget}
Since $\theta^{(0)} = \theta_0$ makes the penalty in
\eqref{eq:regularized} vanish, $J_k(\theta_0) = \loss_k(\theta_0)$.
Consequently, any $\theta$ with $J_k(\theta) \le \loss_k(\theta_0)$
satisfies
\[
\norm{\theta - \theta_0}_{\sal_0}^2
\;\le\; \tfrac{2}{\mu}\bigl(\loss_k(\theta_0) - \loss_k(\theta)\bigr),
\]
and, since the penalty is nonnegative,
$\loss_k(\theta) \le \loss_k(\theta_0)$.
\end{lemma}

\begin{proof}
Writing $J_k(\theta) = \loss_k(\theta) + \tfrac{\mu}{2}
\norm{\theta - \theta_0}_{\sal_0}^2 \le \loss_k(\theta_0)$ and
rearranging gives the first claim; dropping the nonnegative penalty
term gives the second.
\end{proof}

\begin{proof}[Proof of Theorem~\ref{thm:dynamics}(iii)]
By hypothesis $J_k(\hat\theta_k) \le J_k(\theta_0) =
\loss_k(\theta_0)$, so Lemma~\ref{lem:budget} applies directly. The
qualification in the statement is justified because both dynamics of
part (ii) produce monotonically non-increasing values of $J_k$ from
$\theta_0$ (established in the proof of (ii) below), so their limit
points satisfy the hypothesis.
\end{proof}

\begin{proof}[Proof of Theorem~\ref{thm:dynamics}(ii)]
\emph{Deterministic dynamics.} Along the gradient flow
$\dot\theta = -\nabla J_k(\theta)$,
$\tfrac{d}{dt} J_k(\theta) = -\norm{\nabla J_k(\theta)}^2 \le 0$,
so $J_k$ is non-increasing and $J_k(\theta^{(t)}) \le J_k(\theta_0)
= \loss_k(\theta_0)$ for all $t \ge 0$. For gradient descent with
$\gamma \le 1/L_\mu$, the descent lemma applied to the $L_\mu$-smooth
$J_k$ (Lemma~\ref{lem:smoothJ}) gives
$J_k(\theta^{(t+1)}) \le J_k(\theta^{(t)}) -
\gamma(1 - \tfrac{L_\mu\gamma}{2})\norm{\nabla J_k(\theta^{(t)})}^2
\le J_k(\theta^{(t)})$, since $1 - \tfrac{L_\mu\gamma}{2} \ge
\tfrac12 > 0$; hence again $J_k(\theta^{(t)}) \le \loss_k(\theta_0)$.
In either case Lemma~\ref{lem:budget} yields
$\Phi(\theta^{(t)}) \le \tfrac{2}{\mu}(\loss_k(\theta_0) -
\loss_k(\theta^{(t)}))$, and $\loss_k(\theta^{(t)}) \ge \loss_k^\star$
gives the second inequality.

\emph{Stochastic dynamics.} Recall from the proof of part (i)
(Appendix~\ref{app:sgd}, Step~1) the one-step inequality
\[
\E\, J_k(\theta^{(t+1)})
\le \E\, J_k(\theta^{(t)})
- \gamma_T\Bigl(1 - \tfrac{L_\mu\gamma_T}{2}\Bigr)
\E\norm{\nabla J_k(\theta^{(t)})}^2
+ \tfrac{L_\mu\gamma_T^2\sigma^2}{2},
\]
with $\gamma_T = \gamma/\sqrt{T}$. Under the step-size condition
$T \ge \gamma^2 L_\mu^2$ the bracketed coefficient is nonnegative,
so the middle term may be dropped:
$\E J_k(\theta^{(t+1)}) \le \E J_k(\theta^{(t)}) +
\tfrac{L_\mu\gamma_T^2\sigma^2}{2}$. Telescoping from $0$ to
$t \le T$ and using $J_k(\theta_0) = \loss_k(\theta_0)$,
\[
\E\, J_k(\theta^{(t)})
\le \loss_k(\theta_0) + \frac{t\, L_\mu\gamma_T^2\sigma^2}{2}
\le \loss_k(\theta_0) + \frac{L_\mu\gamma^2\sigma^2}{2},
\]
where the last step uses $t \le T$ and $\gamma_T^2 T = \gamma^2$.
Since $\E J_k(\theta^{(t)}) = \E\loss_k(\theta^{(t)}) +
\tfrac{\mu}{2}\E\Phi(\theta^{(t)})$, rearranging gives
\[
\E\,\Phi(\theta^{(t)})
\le \frac{2}{\mu}\bigl(\loss_k(\theta_0) -
\E\loss_k(\theta^{(t)})\bigr) + \frac{\gamma^2 L_\mu\sigma^2}{\mu},
\]
the stated bound, with inflation $\gamma^2 L_\mu\sigma^2/\mu \to 0$
as $\gamma \to 0$. The inflation does not depend on $T$; it is small only for small $\gamma$.
\end{proof}

\subsection{Proof of Theorem~\ref{thm:ceiling}}\label{app:ceiling}

The proof combines three ingredients: the per-task budget of
Theorem~\ref{thm:dynamics}(iii), the Loewner comparison of
Assumption~\ref{ass:domination}, and the triangle inequality in the
$\sal_0$ seminorm. We record the second and third as separate steps
because they are where the metric changes.

\begin{lemma}[Metric transfer]\label{lem:transfer}
Under Assumption~\ref{ass:domination}, for every $k \in [K]$ and every
$x \in \R^d$,
\[
\norm{x}_{\Hess_k}^2 \;\le\; c\,\norm{x}_{\sal_0}^2 .
\]
No definiteness of $\Hess_k$ is required.
\end{lemma}

\begin{proof}
$\Hess_k \preceq c\,\sal_0$ means $c\,\sal_0 - \Hess_k \succeq 0$, i.e.
$x^\top (c\,\sal_0 - \Hess_k) x \ge 0$ for every $x$, which is the
claim after rearranging. The argument uses only that the Loewner order
is defined by nonnegativity of the associated quadratic form, so
$\Hess_k$ may be indefinite; positive semi-definiteness of $\sal_0$,
which \emph{is} assumed (Section~\ref{sec:dynamics}), is what makes
$\norm{\cdot}_{\sal_0}$ a seminorm and is used only in
Lemma~\ref{lem:triangle}.
\end{proof}

\begin{lemma}[Seminorm bound on the displacement]\label{lem:triangle}
Let $\sal_0 \succeq 0$. For every $k \in [K]$ and every
$\bm{\lambda} \in \Lambda \subset \R_{\ge 0}^K$,
\[
\norm{d_k(\bm{\lambda})}_{\sal_0}
\;\le\;
|\lambda_k - 1|\,\norm{\tau_k}_{\sal_0}
\;+\; \sum_{j \ne k} \lambda_j \,\norm{\tau_j}_{\sal_0} .
\]
\end{lemma}

\begin{proof}
Since $\sal_0 \succeq 0$, the map $x \mapsto \norm{x}_{\sal_0} =
(x^\top \sal_0 x)^{1/2}$ is a seminorm: it is absolutely homogeneous,
and the triangle inequality follows from the Cauchy--Schwarz inequality
for the positive semi-definite bilinear form $(x,y) \mapsto x^\top
\sal_0 y$. Applying it to the decomposition
\eqref{eq:displacement},
$d_k(\bm{\lambda}) = (\lambda_k - 1)\tau_k + \sum_{j \ne k}
\lambda_j \tau_j$, and using $\lambda_j \ge 0$ for $j \ne k$ gives the
claim.
\end{proof}

\begin{proof}[Proof of Theorem~\ref{thm:ceiling}]
\emph{Step 1: per-task budgets.} Fix $j \in [K]$. By hypothesis the
task-$j$ run satisfies $J_j(\hat\theta_j) \le J_j(\theta_0)$, and since
training initializes at $\theta_0$ the penalty in \eqref{eq:regularized}
vanishes there, so $J_j(\theta_0) = \loss_j(\theta_0)$. Lemma~\ref{lem:budget}
therefore applies and gives
\begin{equation}\label{eq:perbudget}
\norm{\tau_j}_{\sal_0}^2
\;=\; \norm{\hat\theta_j - \theta_0}_{\sal_0}^2
\;\le\; \frac{2}{\mu}\bigl(\loss_j(\theta_0) - \loss_j(\hat\theta_j)\bigr)
\;=\; \frac{2\Delta_j}{\mu} ,
\end{equation}
which is Theorem~\ref{thm:dynamics}(iii). Note $\Delta_j \ge 0$ by the
second conclusion of Lemma~\ref{lem:budget}, so the square root below is
well defined. This step is where the hypothesis that every task is
trained under the \emph{same} $\sal_0$ and the \emph{same} $\mu$ is
used: the budgets \eqref{eq:perbudget} are then all expressed in one
common metric and one common scale, which is what permits them to be
combined in Step~2.

\emph{Step 2: combination in the common metric.} By
Lemma~\ref{lem:triangle} and \eqref{eq:perbudget},
\[
\norm{d_k(\bm{\lambda})}_{\sal_0}
\;\le\;
|\lambda_k - 1|\sqrt{\frac{2\Delta_k}{\mu}}
+ \sum_{j \ne k}\lambda_j \sqrt{\frac{2\Delta_j}{\mu}}
\;=\;
\sqrt{\frac{2}{\mu}}
\Bigl(
|\lambda_k - 1|\sqrt{\Delta_k} + \sum_{j \ne k}\lambda_j\sqrt{\Delta_j}
\Bigr).
\]

\emph{Step 3: transfer to the task metric.} By Lemma~\ref{lem:identity},
$\Ov_k(\bm{\lambda}) = \norm{d_k(\bm{\lambda})}_{\Hess_k}^2$, and by
Lemma~\ref{lem:transfer} this is at most
$c \norm{d_k(\bm{\lambda})}_{\sal_0}^2$. Squaring the display of
Step~2 and multiplying by $c$ gives \eqref{eq:ceiling}.

\emph{Simplified form.} With $\Delta = \max_j \Delta_j$, the bracket in
\eqref{eq:ceiling} is at most
$\bigl(|\lambda_k - 1| + \sum_{j\ne k}\lambda_j\bigr)\sqrt{\Delta}
= \norm{\bm{\lambda} - e_k}_1 \sqrt{\Delta}$, the equality because
$\lambda_j \ge 0$ for $j \ne k$; substituting and maximizing over $k$
gives \eqref{eq:ceiling-simple}.

\emph{Composability consequence.} By Theorem~\ref{thm:bound},
$\loss_k(\merged) - \loss_k(\theta_k) \le \tfrac12 \Ov_k(\bm{\lambda})
+ \epsilon_g\norm{d_k} + \tfrac{\rho}{6}\norm{d_k}^3$ for every $k$.
Substituting \eqref{eq:ceiling} for $\tfrac12\Ov_k$ and requiring the
result to be at most $\varepsilon$ for every $k$ is
Definition~\ref{def:composability}.
\end{proof}

\begin{remark}[Where the bound is loose]\label{rem:ceilingloose}
Two of the three steps are worst-case. Lemma~\ref{lem:transfer}
replaces the task curvature by its uniform upper comparison and is
tight only in the directions attaining $c$; Lemma~\ref{lem:triangle}
assumes the task vectors are aligned in the $\sal_0$ seminorm and is
tight only when they are positively parallel. The second is the more
consequential: the whole content of Section~\ref{sec:bound} is that
interference depends on the \emph{signed} alignment between task
vectors, and the triangle inequality of Step~2 discards exactly that
sign, assuming the adversarial case in which every cross-task overlap
term of \eqref{eq:omega} is maximally positive. Theorem~\ref{thm:ceiling}
is thus a guarantee that holds for any sibling collection produced this
way, including the worst one, and is correspondingly conservative for
typical collections---which is the appropriate posture for a bound whose
purpose is to certify before the siblings are known. Recovering the
alignment structure would require joint control of $\tau_k^\top \sal_0
\tau_j$ across tasks, which the single-task training procedure of
Algorithm~\ref{alg:mat} cannot supply by construction, since it never
sees a sibling.
\end{remark}

\subsection{Proof of Corollary~\ref{cor:factorized}}\label{app:factorized}

We use two standard facts, recorded for completeness.

\begin{lemma}[Quadratic form perturbation]\label{lem:quadpert}
For symmetric matrices $A, B$ and any $x$,
$\bigl| \norm{x}_A^2 - \norm{x}_B^2 \bigr|
\le \norm{A - B}_{\mathrm{op}} \norm{x}^2$.
\end{lemma}

\begin{proof}
$\norm{x}_A^2 - \norm{x}_B^2 = x^\top (A - B)\, x$, and
$|x^\top M x| \le \norm{M}_{\mathrm{op}} \norm{x}^2$ for symmetric
$M$.
\end{proof}

\begin{lemma}[Truncation residual]\label{lem:trunc}
Let $\sal \succeq 0$ be symmetric with spectral decomposition
$\sal = \sum_i \sigma_i u_i u_i^\top$, $\sigma_1 \ge \sigma_2 \ge
\cdots \ge 0$, and $\sal^{(r)} = \sum_{i \le r} \sigma_i u_i
u_i^\top$. Then $\sal^{(r)}$ is symmetric,
$\sal - \sal^{(r)} \succeq 0$,
$\norm{\sal - \sal^{(r)}}_{\mathrm{op}} = \sigma_{r+1} =
\epsilon_r(\sal)$, $\sal^{(r)} \succeq 0$, and
$\norm{\sal^{(r)}}_{\mathrm{op}} = \sigma_1 =
\norm{\sal}_{\mathrm{op}}$ for every $r \ge 1$.
\end{lemma}

\begin{proof}
All four claims are immediate from the decomposition:
$\sal - \sal^{(r)} = \sum_{i > r} \sigma_i u_i u_i^\top$ is a
nonnegative combination of rank-one projections with largest
coefficient $\sigma_{r+1}$, and $\sal^{(r)}$ retains the top
eigenvalue $\sigma_1$.
\end{proof}

\begin{proof}[Proof of Corollary~\ref{cor:factorized}]
\emph{(a)} By the triangle inequality,
$\norm{\sal_k^{(r)} - \Hess_k}_{\mathrm{op}} \le
\norm{\sal_k^{(r)} - \sal_k}_{\mathrm{op}} +
\norm{\sal_k - \Hess_k}_{\mathrm{op}} \le \epsilon_r(\sal_k) +
\delta$, using Lemma~\ref{lem:trunc} and
Assumption~\ref{ass:saliency}. Lemma~\ref{lem:quadpert} with
$A = \sal_k^{(r)}$, $B = \Hess_k$, $x = d_k(\bm{\lambda})$, and
Lemma~\ref{lem:identity} then give
\[
\bigl|\widehat{\Ov}_k - \Ov_k(\bm{\lambda})\bigr|
\;\le\; \bigl(\delta + \epsilon_r(\sal_k)\bigr)
\norm{d_k(\bm{\lambda})}^2 .
\]
Substituting $\Ov_k \le \widehat{\Ov}_k + (\delta +
\epsilon_r(\sal_k))\norm{d_k}^2$ into Theorem~\ref{thm:bound} yields
the claimed inflated bound. For the last claim, Weyl's inequality
for symmetric matrices and Assumption~\ref{ass:saliency} give
$\sigma_{r+1}(\sal_k) \le \sigma_{r+1}(\Hess_k) + \delta$, while
Assumption~\ref{ass:rank} gives
$\sigma_{r+1}(\Hess_k) \le \sum_{i > r} \sigma_i(\Hess_k) \le \eta
\sum_{i \ge 1} \sigma_i(\Hess_k) = \eta \operatorname{tr}(\Hess_k)$,
the last equality since singular values coincide with eigenvalues
for $\Hess_k \succeq 0$; for indefinite $\Hess_k$ the same bound
holds with $\operatorname{tr}(\Hess_k)$ replaced by the trace norm
$\norm{\Hess_k}_*$. The alternative branch of the minimum follows
without Assumption~\ref{ass:rank}: for $\Hess_k \succeq 0$ the top
$r+1$ eigenvalues are each at least $\sigma_{r+1}(\Hess_k)$ and sum
to at most $\operatorname{tr}(\Hess_k)$, so
$\sigma_{r+1}(\Hess_k) \le \operatorname{tr}(\Hess_k)/(r+1)$.

\emph{(b)} By Lemma~\ref{lem:trunc}, $\sal_0^{(r)}$ is symmetric
positive semi-definite with
$\norm{\sal_0^{(r)}}_{\mathrm{op}} = \norm{\sal_0}_{\mathrm{op}}$.
The proofs of Theorem~\ref{thm:dynamics}(i)
(Appendix~\ref{app:sgd}) and (ii)--(iii) (Appendix~\ref{app:budget},
including the rearrangement of Lemma~\ref{lem:budget}) use
only three properties of the regularizer matrix: symmetry, consumed
in the gradient formula
$\nabla\bigl[\tfrac{\mu}{2}\norm{\theta - \theta_0}_{S}^2\bigr]
= \mu S(\theta - \theta_0)$; positive semi-definiteness
(Step~4, Lemma~\ref{lem:budget}, and dropping the penalty in the
descent arguments); and the operator norm (Lemma~\ref{lem:smoothJ},
which fixes $L_\mu$ in the step-size conditions of (i) and (ii)).
All three are supplied by Lemma~\ref{lem:trunc}, so all three proofs
apply verbatim with $\sal_0^{(r)}$ in place of $\sal_0$, and the
constant $L_\mu$ in parts (i) and (ii) is unchanged.

\emph{(c)} Write $x = \hat\theta_k - \theta_0$. By part (b),
Theorem~\ref{thm:dynamics}(iii) applied to the objective
$J_k^{(r)}$ gives $\norm{x}_{\sal_0^{(r)}}^2 \le
\tfrac{2}{\mu}(\loss_k(\theta_0) - \loss_k(\hat\theta_k))$. By
Lemma~\ref{lem:trunc}, $\sal_0 - \sal_0^{(r)} \succeq 0$ with
operator norm $\epsilon_r(\sal_0)$, so
\[
\norm{x}_{\sal_0}^2
= \norm{x}_{\sal_0^{(r)}}^2 + x^\top (\sal_0 - \sal_0^{(r)})\, x
\;\le\; \norm{x}_{\sal_0^{(r)}}^2 + \epsilon_r(\sal_0)\norm{x}^2 ,
\]
and combining the two displays gives the claim. (The upper bound
uses only the operator norm of the residual; positive
semi-definiteness additionally implies
$\norm{x}_{\sal_0^{(r)}}^2 \le \norm{x}_{\sal_0}^2$, so the
truncated budget always \emph{understates} the true one---the
conservative direction.)
\end{proof}

\begin{remark}[What is and is not being claimed]\label{rem:factremark}
Part (b) is exact, not approximate: training with the truncated
statistic loses nothing in the convergence guarantee and produces
its own budget in the truncated metric. The approximation enters
only when results are read in a metric other than the one trained
with---part (a) at merge time, part (c) for the full-metric
budget---and there the error is the measurable spectral tail
$\epsilon_r$, consistent with the empirical finding of
\citep{moayedikia2026bridging} that low-rank truncation of
trajectory statistics preserves merging-relevant information. One
scope caveat: the corollary is stated for spectral truncation,
whereas practical memory-efficient optimizers may maintain other
factorizations (e.g.\ Adafactor's nonnegative row--column
factorization of a diagonal). Parts (a) and (c) hold for any
symmetric approximation with $\epsilon_r$ replaced by
$\norm{\sal - \sal^{(r)}}_{\mathrm{op}}$, and part (b) holds for
any symmetric positive semi-definite approximation with
$\norm{\sal_0^{(r)}}_{\mathrm{op}}$ in $L_\mu$; only the
``constant unchanged'' clause is specific to spectral truncation.
\end{remark}

\section*{Data availability}
The code for the three numerical studies and their stored results are
provided as ancillary files with the arXiv version of this paper. The
datasets used (GLUE, WikiText-2) are publicly available.

\bibliographystyle{abbrvnat}
\bibliography{paperA_refs}

@inproceedings{ilharco2023editing,
  author    = {Gabriel Ilharco and Marco Tulio Ribeiro and Mitchell Wortsman and Suchin Gururangan and Ludwig Schmidt and Hannaneh Hajishirzi and Ali Farhadi},
  title     = {Editing Models with Task Arithmetic},
  booktitle = {International Conference on Learning Representations (ICLR)},
  year      = {2023}
}

@inproceedings{ortizjimenez2023task,
  author    = {Guillermo Ortiz-Jimenez and Alessandro Favero and Pascal Frossard},
  title     = {Task Arithmetic in the Tangent Space: Improved Editing of Pre-Trained Models},
  booktitle = {Advances in Neural Information Processing Systems (NeurIPS)},
  volume    = {36},
  year      = {2023}
}

@inproceedings{matena2022fisher,
  author    = {Michael S. Matena and Colin A. Raffel},
  title     = {Merging Models with Fisher-Weighted Averaging},
  booktitle = {Advances in Neural Information Processing Systems (NeurIPS)},
  volume    = {35},
  pages     = {17703--17716},
  year      = {2022}
}

@inproceedings{yadav2023ties,
  author    = {Prateek Yadav and Derek Tam and Leshem Choshen and Colin Raffel and Mohit Bansal},
  title     = {{TIES}-Merging: Resolving Interference When Merging Models},
  booktitle = {Advances in Neural Information Processing Systems (NeurIPS)},
  volume    = {36},
  year      = {2023}
}

@inproceedings{yu2024dare,
  author    = {Le Yu and Bowen Yu and Haiyang Yu and Fei Huang and Yongbin Li},
  title     = {Language Models are Super Mario: Absorbing Abilities from Homologous Models as a Free Lunch},
  booktitle = {Proceedings of the 41st International Conference on Machine Learning (ICML)},
  series    = {Proceedings of Machine Learning Research},
  volume    = {235},
  year      = {2024}
}

@inproceedings{li2018explicit,
  author    = {Xuhong Li and Yves Grandvalet and Franck Davoine},
  title     = {Explicit Inductive Bias for Transfer Learning with Convolutional Networks},
  booktitle = {Proceedings of the 35th International Conference on Machine Learning (ICML)},
  series    = {Proceedings of Machine Learning Research},
  volume    = {80},
  pages     = {2825--2834},
  year      = {2018}
}

@article{kirkpatrick2017overcoming,
  author    = {James Kirkpatrick and Razvan Pascanu and Neil Rabinowitz and Joel Veness and Guillaume Desjardins and Andrei A. Rusu and Kieran Milan and John Quan and Tiago Ramalho and Agnieszka Grabska-Barwinska and Demis Hassabis and Claudia Clopath and Dharshan Kumaran and Raia Hadsell},
  title     = {Overcoming Catastrophic Forgetting in Neural Networks},
  journal   = {Proceedings of the National Academy of Sciences},
  volume    = {114},
  number    = {13},
  pages     = {3521--3526},
  year      = {2017}
}

@article{barakat2021convergence,
  author    = {Anas Barakat and Pascal Bianchi},
  title     = {Convergence and Dynamical Behavior of the {ADAM} Algorithm for Nonconvex Stochastic Optimization},
  journal   = {SIAM Journal on Optimization},
  volume    = {31},
  number    = {1},
  pages     = {244--274},
  year      = {2021}
}

@inproceedings{porrello2026dataless,
  author    = {Angelo Porrello and Pietro Buzzega and Felix Dangel and Thomas Sommariva and Riccardo Salami and Lorenzo Bonicelli and Simone Calderara},
  title     = {Dataless Weight Disentanglement in Task Arithmetic via Kronecker-Factored Approximate Curvature},
  booktitle = {International Conference on Learning Representations (ICLR)},
  year      = {2026}
}

@article{moayedikia2026bridging,
  author    = {Alireza Moayedikia and Alicia Troncoso},
  title     = {Bridging Training and Merging Through Momentum-Aware Optimization},
  journal   = {Information Sciences},
  volume    = {745},
  pages     = {123402},
  year      = {2026}
}

@article{yang2026model,
  author    = {Enneng Yang and Li Shen and Guibing Guo and Xingwei Wang and Xiaochun Cao and Jie Zhang and Dacheng Tao},
  title     = {Model Merging in {LLMs}, {MLLMs}, and Beyond: Methods, Theories, Applications and Opportunities},
  journal   = {ACM Computing Surveys},
  year      = {2026}
}

@misc{lu2026geometry,
  author    = {Juanwu Lu and Anand Bhaskar and Brian Axelrod and Ekaterina Tolstaya and Tristan Emrich},
  title     = {Model Merging on Loss Landscape: A Geometry Perspective},
  howpublished = {arXiv preprint arXiv:2605.26693},
  year      = {2026}
}

@misc{zhou2024atm,
  author    = {Luca Zhou and Daniele Solombrino and Donato Crisostomi and Maria Sofia Bucarelli and Fabrizio Silvestri and Emanuele Rodol{\`a}},
  title     = {{ATM}: Improving Model Merging by Alternating Tuning and Merging},
  howpublished = {arXiv preprint arXiv:2411.03055},
  year      = {2024}
}

@inproceedings{demystifying2026,
  author    = {Luca Zhou and Bo Zhao and Rose Yu and Emanuele Rodol{\`a}},
  title     = {Demystifying Mergeability: Interpretable Properties to Predict Model Merging Success},
  booktitle = {Proceedings of the 43rd International Conference on Machine Learning (ICML)},
  year      = {2026}
}

@misc{li2026unified,
  author    = {Qinglun Li and Anke Tang and Miao Zhang and Mengzhu Wang and Quanjun Yin and Li Shen},
  title     = {A Unified Generalization Framework for Model Merging: Trade-offs, Non-Linearity, and Scaling Laws},
  howpublished = {arXiv preprint arXiv:2601.21690},
  year      = {2026}
}

@article{ghadimi2013stochastic,
  author  = {Saeed Ghadimi and Guanghui Lan},
  title   = {Stochastic First- and Zeroth-Order Methods for Nonconvex Stochastic Programming},
  journal = {SIAM Journal on Optimization},
  volume  = {23}, number = {4}, pages = {2341--2368}, year = {2013}
}

@article{bottou2018optimization,
  author  = {L{\'e}on Bottou and Frank E. Curtis and Jorge Nocedal},
  title   = {Optimization Methods for Large-Scale Machine Learning},
  journal = {SIAM Review}, volume = {60}, number = {2}, pages = {223--311}, year = {2018}
}

@inproceedings{kingma2015adam,
  author    = {Diederik P. Kingma and Jimmy Ba},
  title     = {Adam: A Method for Stochastic Optimization},
  booktitle = {International Conference on Learning Representations (ICLR)}, year = {2015}
}

@inproceedings{loshchilov2019decoupled,
  author    = {Ilya Loshchilov and Frank Hutter},
  title     = {Decoupled Weight Decay Regularization},
  booktitle = {International Conference on Learning Representations (ICLR)}, year = {2019}
}

@inproceedings{reddi2018convergence,
  author    = {Sashank J. Reddi and Satyen Kale and Sanjiv Kumar},
  title     = {On the Convergence of {Adam} and Beyond},
  booktitle = {International Conference on Learning Representations (ICLR)}, year = {2018}
}

@article{defossez2022simple,
  author  = {Alexandre D{\'e}fossez and L{\'e}on Bottou and Francis Bach and Nicolas Usunier},
  title   = {A Simple Convergence Proof of {Adam} and {Adagrad}},
  journal = {Transactions on Machine Learning Research (TMLR)}, year = {2022}
}

@inproceedings{karimi2016linear,
  author    = {Hamed Karimi and Julie Nutini and Mark Schmidt},
  title     = {Linear Convergence of Gradient and Proximal-Gradient Methods Under the {Polyak}--{\L}ojasiewicz Condition},
  booktitle = {European Conference on Machine Learning and Knowledge Discovery in Databases (ECML-PKDD)}, year = {2016}
}

@inproceedings{shazeer2018adafactor,
  author    = {Noam Shazeer and Mitchell Stern},
  title     = {Adafactor: Adaptive Learning Rates with Sublinear Memory Cost},
  booktitle = {Proceedings of the 35th International Conference on Machine Learning (ICML)}, year = {2018}
}

@inproceedings{sagun2018empirical,
  author    = {Levent Sagun and Utku Evci and V. Ugur G{\"u}ney and Yann Dauphin and L{\'e}on Bottou},
  title     = {Empirical Analysis of the {Hessian} of Over-Parametrized Neural Networks},
  booktitle = {International Conference on Learning Representations (ICLR) Workshop}, year = {2018}
}

@misc{gurari2018gradient,
  author       = {Guy Gur-Ari and Daniel A. Roberts and Ethan Dyer},
  title        = {Gradient Descent Happens in a Tiny Subspace},
  howpublished = {arXiv preprint arXiv:1812.04754}, year = {2018}
}

@inproceedings{ghorbani2019investigation,
  author    = {Behrooz Ghorbani and Shankar Krishnan and Ying Xiao},
  title     = {An Investigation into Neural Net Optimization via {Hessian} Eigenvalue Density},
  booktitle = {Proceedings of the 36th International Conference on Machine Learning (ICML)}, year = {2019}
}

@article{papyan2020traces,
  author  = {Vardan Papyan},
  title   = {Traces of Class/Cross-Class Structure Pervade Deep Learning Spectra},
  journal = {Journal of Machine Learning Research}, volume = {21}, pages = {1--64}, year = {2020}
}

@inproceedings{wortsman2022soups,
  author    = {Mitchell Wortsman and Gabriel Ilharco and Samir Yitzhak Gadre and others},
  title     = {Model Soups: Averaging Weights of Multiple Fine-Tuned Models Improves Accuracy Without Increasing Inference Time},
  booktitle = {Proceedings of the 39th International Conference on Machine Learning (ICML)}, year = {2022}
}

@inproceedings{izmailov2018averaging,
  author    = {Pavel Izmailov and Dmitrii Podoprikhin and Timur Garipov and Dmitry Vetrov and Andrew Gordon Wilson},
  title     = {Averaging Weights Leads to Wider Optima and Better Generalization},
  booktitle = {Conference on Uncertainty in Artificial Intelligence (UAI)}, year = {2018}
}

@inproceedings{frankle2020linear,
  author    = {Jonathan Frankle and Gintare Karolina Dziugaite and Daniel M. Roy and Michael Carbin},
  title     = {Linear Mode Connectivity and the Lottery Ticket Hypothesis},
  booktitle = {Proceedings of the 37th International Conference on Machine Learning (ICML)}, year = {2020}
}

@inproceedings{entezari2022role,
  author    = {Rahim Entezari and Hanie Sedghi and Olga Saukh and Behnam Neyshabur},
  title     = {The Role of Permutation Invariance in Linear Mode Connectivity of Neural Networks},
  booktitle = {International Conference on Learning Representations (ICLR)}, year = {2022}
}

@inproceedings{ainsworth2023gitrebasin,
  author    = {Samuel K. Ainsworth and Jonathan Hayase and Siddhartha Srinivasa},
  title     = {Git Re-Basin: Merging Models Modulo Permutation Symmetries},
  booktitle = {International Conference on Learning Representations (ICLR)}, year = {2023}
}

@inproceedings{jin2023regmean,
  author    = {Xisen Jin and Xiang Ren and Daniel Preo{\c{t}}iuc-Pietro and Pengxiang Cheng},
  title     = {Dataless Knowledge Fusion by Merging Weights of Language Models},
  booktitle = {International Conference on Learning Representations (ICLR)}, year = {2023}
}

@inproceedings{yang2024adamerging,
  author    = {Enneng Yang and Zhenyi Wang and Li Shen and others},
  title     = {{AdaMerging}: Adaptive Model Merging for Multi-Task Learning},
  booktitle = {International Conference on Learning Representations (ICLR)}, year = {2024}
}

@inproceedings{stoica2024zipit,
  author    = {George Stoica and Daniel Bolya and Jakob Bjorner and Pratik Ramesh and Taylor Hearn and Judy Hoffman},
  title     = {{ZipIt!} Merging Models from Different Tasks Without Training},
  booktitle = {International Conference on Learning Representations (ICLR)}, year = {2024}
}

@inproceedings{martens2015kfac,
  author    = {James Martens and Roger Grosse},
  title     = {Optimizing Neural Networks with Kronecker-Factored Approximate Curvature},
  booktitle = {Proceedings of the 32nd International Conference on Machine Learning (ICML)}, year = {2015}
}

@inproceedings{kunstner2019limitations,
  author    = {Frederik Kunstner and Philipp Hennig and Lukas Balles},
  title     = {Limitations of the Empirical {Fisher} Approximation for Natural Gradient Descent},
  booktitle = {Advances in Neural Information Processing Systems (NeurIPS)}, year = {2019}
}

@article{amari1998natural,
  author  = {Shun-Ichi Amari},
  title   = {Natural Gradient Works Efficiently in Learning},
  journal = {Neural Computation}, volume = {10}, number = {2}, pages = {251--276}, year = {1998}
}

@inproceedings{jacot2018ntk,
  author    = {Arthur Jacot and Franck Gabriel and Cl{\'e}ment Hongler},
  title     = {Neural Tangent Kernel: Convergence and Generalization in Neural Networks},
  booktitle = {Advances in Neural Information Processing Systems (NeurIPS)}, year = {2018}
}

@inproceedings{lee2019wide,
  author    = {Jaehoon Lee and Lechao Xiao and Samuel S. Schoenholz and Yasaman Bahri and Roman Novak and Jascha Sohl-Dickstein and Jeffrey Pennington},
  title     = {Wide Neural Networks of Any Depth Evolve as Linear Models Under Gradient Descent},
  booktitle = {Advances in Neural Information Processing Systems (NeurIPS)}, year = {2019}
}

@inproceedings{zenke2017continual,
  author    = {Friedemann Zenke and Ben Poole and Surya Ganguli},
  title     = {Continual Learning Through Synaptic Intelligence},
  booktitle = {Proceedings of the 34th International Conference on Machine Learning (ICML)}, year = {2017}
}

@inproceedings{aljundi2018memory,
  author    = {Rahaf Aljundi and Francesca Babiloni and Mohamed Elhoseiny and Marcus Rohrbach and Tinne Tuytelaars},
  title     = {Memory Aware Synapses: Learning What (Not) to Forget},
  booktitle = {European Conference on Computer Vision (ECCV)}, year = {2018}
}

@inproceedings{yu2020gradient,
  author    = {Tianhe Yu and Saurabh Kumar and Abhishek Gupta and Sergey Levine and Karol Hausman and Chelsea Finn},
  title     = {Gradient Surgery for Multi-Task Learning},
  booktitle = {Advances in Neural Information Processing Systems (NeurIPS)}, year = {2020}
}

\end{document}